\documentclass{article}

\usepackage{microtype}
\usepackage{graphicx}
\usepackage{subcaption}
\usepackage{booktabs} % for professional tables

\usepackage{hyperref}

\usepackage[preprint]{icml2026}

\usepackage{amsmath}
\usepackage{amssymb}
\usepackage{mathtools}
\usepackage{amsthm}

\usepackage{adjustbox}
\usepackage{booktabs}
\usepackage{multirow}
\usepackage{array}
\usepackage{subcaption}
\usepackage{xcolor}
\usepackage{siunitx}
\usepackage{nicematrix}
\usepackage{natbib}
\usepackage{bm}
\usepackage{amsmath,amssymb}

\usepackage[capitalize,noabbrev]{cleveref}

\theoremstyle{plain}

\theoremstyle{definition}

\theoremstyle{remark}

\newcolumntype{y}[1]{>{\raggedright\arraybackslash}p{#1}}
\newcolumntype{x}[1]{>{\centering\arraybackslash}p{#1}}

\definecolor{AppleGreen}{HTML}{00B050}
\definecolor{MyBlue}{HTML}{98B1DF}

\usepackage[textsize=tiny]{todonotes}

\icmltitlerunning{FlowConsist: Make Your Flow Consistent with Real Trajectory}

\begin{document}

\twocolumn[
  \icmltitle{FlowConsist: Make Your Flow Consistent with Real Trajectory}



  \icmlsetsymbol{equal}{*}

  \begin{icmlauthorlist}
    \icmlauthor{Tianyi Zhang}{1,2} ~~
    \icmlauthor{Chengcheng Liu}{2} ~~
    \icmlauthor{Jinwei Chen}{2} ~~
    \icmlauthor{Chun-Le Guo}{1,3} ~~
    \icmlauthor{Chongyi Li}{1,3} ~~
    \\
    \icmlauthor{Ming-Ming Cheng}{1,3} ~~
    \icmlauthor{Bo Li}{2} ~~
    \icmlauthor{Peng-Tao Jiang}{2} ~~
  \end{icmlauthorlist}

  \icmlaffiliation{1}{VCIP, CS, Nankai University}
  \icmlaffiliation{2}{vivo Mobile Communication Co. Ltd}
  \icmlaffiliation{3}{NKIARI, Shenzhen Futian}

  \icmlcorrespondingauthor{Chun-Le Guo}{guochunle@nankai.edu.cn}
  \icmlcorrespondingauthor{Peng-Tao Jiang}{pt.jiang@vivo.com}
  \icmlkeywords{Machine Learning, ICML}

  \vskip 0.3in
]



\printAffiliationsAndNotice{}  % no special notice (required even if empty)

\begin{abstract}
Fast flow models accelerate the iterative sampling process by learning to directly predict ODE path integrals, enabling one-step or few-step generation. 
However, we argue that current fast-flow training paradigms suffer from two fundamental issues. 
First, conditional velocities constructed from randomly paired noise–data samples introduce systematic trajectory drift, preventing models from following a consistent ODE path. 
Second, the model's approximation errors accumulate over time steps, leading to severe deviations across long time intervals.
To address these issues, we propose FlowConsist, a training framework designed to enforce trajectory consistency in fast flows. 
We propose a principled alternative that replaces conditional velocities with the marginal velocities predicted by the model itself, aligning optimization with the true trajectory.
To further address error accumulation over time steps, we introduce a trajectory rectification strategy that aligns the marginal distributions of generated and real samples at every time step along the trajectory.
Our method establishes a new state-of-the-art on ImageNet 256×256, achieving an FID of 1.52 with only 1 sampling step. 
\end{abstract}

\section{Introduction}

Diffusion models~\cite{diffusion,ddpm,ncsn,scoresde,karras2022elucidating} and flow matching~\cite{rectified,xu2022poisson,lipman2023flow,xu2023pfgm++} have demonstrated remarkable performance in generative modeling~\cite{rombach2022high,watson2023novo}.
However, they generally require a high Number of Function Evaluations (NFE) for synthesis, since integrating the underlying Ordinary or Stochastic Differential Equation (ODE/SDE) from the prior to the data distribution involves many solver steps.

Recent works~\cite{cm,mf,imm,ect,scm,shortcut,boffi2025flow, sun2025anystep, cheng2025twinflow} aim to reduce the number of sampling steps in the diffusion/flow process to improve generation efficiency, which we refer to as fast flow models.
By directly predicting the solution of the ODE, fast flow models achieve a direct mapping between time steps along the ODE trajectory. 
However, these methods generally face two significant limitations: 
(1) Theoretically, the conditional velocity constructed from training data causes the optimization objective to drift away from the intended ODE trajectory.
(2) Practically, approximation errors introduced by the model accumulate significantly over large time spans. 
% (2) Practically, approximation errors introduced from \jpt{the iterative sampling process} accumulate significantly over large time spans. 

\begin{figure}
    \centering
    \includegraphics[width=1.0\linewidth]{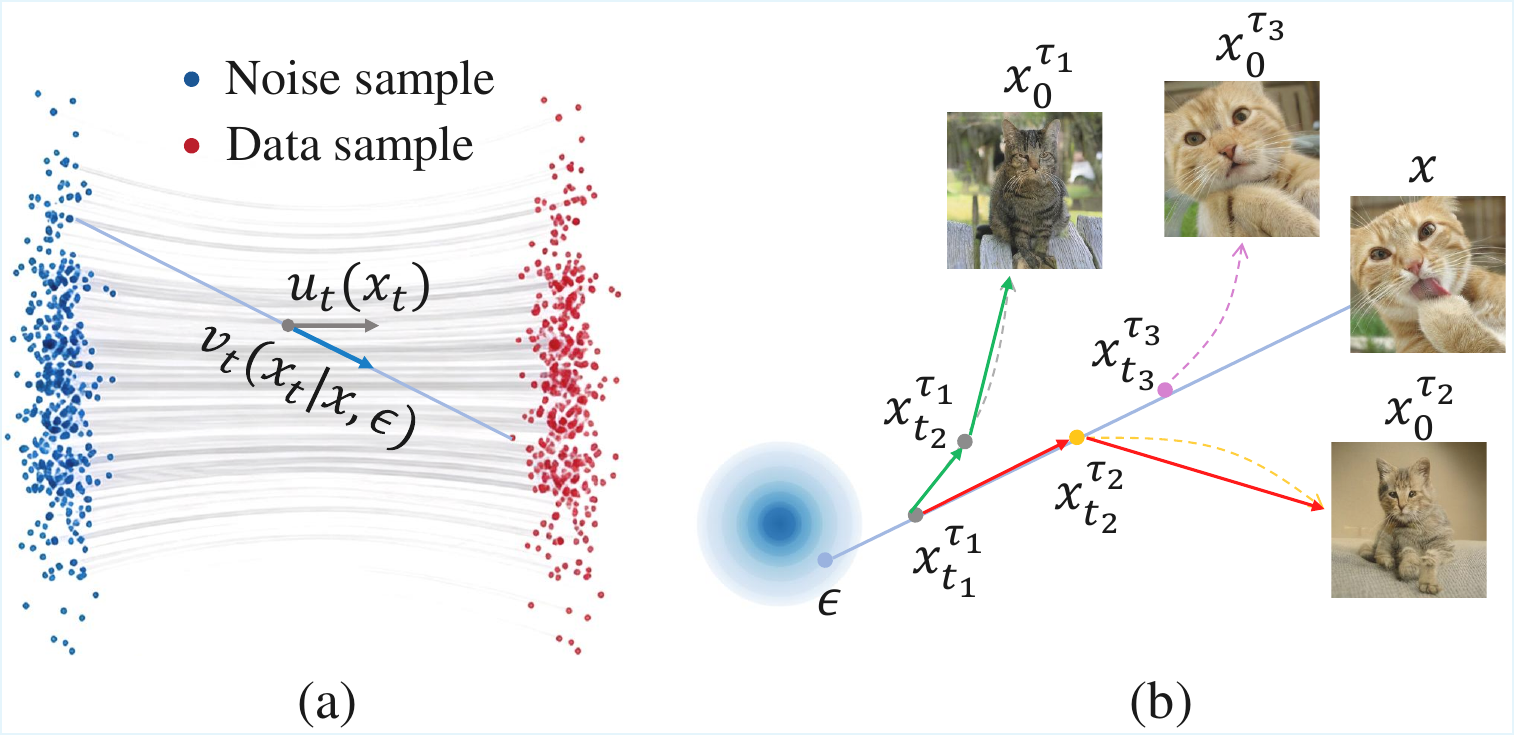}
    \caption{(a) Marginal flow fields and conditional paths. A conditional path (\textcolor{MyBlue}{blue line}) and its conditional velocity $v_t(x_t|x, \epsilon)$ can be constructed by pairing a randomly sampled noise $\epsilon$ with a data $x$. By marginalizing over all possible conditioned paths, we obtain a marginal flow field (\textcolor{gray}{gray lines}). When constructing a randomly paired conditional path, it becomes evident that such a path intersects multiple distinct marginal path. (b) We construct a conditional path (\textcolor{MyBlue}{blue line}) by perturbing the image $x_0$ with randomly sampled noise $\epsilon$. We initialize the marginal paths (dashed lines using a pre-trained flow model) from various time steps along this conditional path and obtain distinct data samples. This demonstrates that the conditional velocity induces a drift in the marginal trajectories, causing them to deviate from the ground-truth trajectory (\textcolor{AppleGreen}{green}) toward erroneous trajectories (\textcolor{red}{red}).}
    \label{fig:teaser}
    % \vspace{-1em}
\end{figure}

We first analyze trajectory drift from the perspective of Flow Matching (FM).
FM constructs conditional velocities based on randomly sampled noise-data pairs and trains the model to recover the marginal velocity field by learning the expectation over all possible conditional velocities. 
Fast flow models, such as Consistency Model~\cite{cm,scm}, MeanFlow~\cite{mf} and their variants~\cite{geng2025improved,you2025modular,tong2025flow, sun2025anystep}, aim to establish a direct mapping between time steps by learning the average velocity of this marginal field. 
However, due to infeasible integration and inaccessible marginal velocity during training, these methods typically reformulate the training objective in differential form and employ conditional velocity as a surrogate for the marginal velocity.

Although the conditional velocity equals the marginal velocity in expectation, we point out that reliance on such conditional velocities leads to a systematic drift in the optimized trajectories.
As illustrated in Fig.~\ref{fig:teaser}(a), a conditional path constructed from a random noise–data pair intersects multiple marginal trajectories.
This implies that moving along the conditional velocity causes the current marginal trajectory to shift onto a different one. 
We empirically validate this in Fig.~\ref{fig:teaser}(b), tracing marginal trajectories (estimated by a pre-trained flow model) from different points along a single conditional path leads to divergent results.
Therefore, training a fast flow model along conditional velocities shifts the optimization target from the ground-truth marginal trajectory (green) to a different one (red).
Such trajectory drift theoretically prevents fast flow models from following a consistent, singular trajectory, thereby degrading their performance.
To rectify this, we propose replacing randomly sampled conditional velocities with the self-predicted marginal velocities of the model. 
By aligning the flow with its authentic marginal field instead of individual noise-data pairs, our approach ensures trajectory consistency.

In addition, the training objectives of fast flow models are derived from self-predicted velocities, which leads to a continuous accumulation of approximation errors over large time spans. 
This hinders high-fidelity mappings from noise to data.
To manifest these cumulative errors, we propose a trajectory rectification strategy. 
By aligning the marginal distribution of model-generated samples with that of real data, we rectify the mappings from all points along the ODE trajectory back to the data distribution. 
Unlike existing methods that apply such corrections only to the initial noise~\cite{dmd,dmd2,tong2025flow}, our approach extends this rectification to every time step along the trajectory, ensuring seamless integration with the trajectory consistency objective. 
In practice, this rectification is fully teacher-free, as we utilize the self-predicted marginal velocities of the model throughout. 
We term our method FlowConsist, as it is designed with the core mission of achieving fast flows that remain inherently consistent with their trajectories.

To validate the efficacy of our method, we train FlowConsist on ImageNet 256$\times$256, which achieves a state-of-the-art (SOTA) FID of 1.52 with 1-NFE.
Remarkably, our method achieves substantial performance gains solely by refining the training objective, underscoring the importance of trajectory consistency for fast flow models. 
We believe that FlowConsist offers a new perspective for the development of the fast flow modeling.

\section{Related Work}
\textbf{Diffusion and Flow Matching.} 
Diffusion models~\cite{diffusion,ddpm,ncsn,scoresde,karras2022elucidating} and flow matching~\cite{rectified,xu2022poisson,lipman2023flow,xu2023pfgm++} are two closely related paradigms for learning continuous generative dynamics. 
These approaches formulate generative modeling as transporting probability distributions via stochastic or ordinary differential equations (SDEs/ODEs) trajectory. 
Typically, a neural network is trained to learn the score function or the velocity field along this trajectory, and sampling is performed by numerically integrating the corresponding ODE or SDE.

\textbf{Fast Flow Generative Models.} Considering that diffusion and flow matching require multi-step numerical integration of ODEs/SDEs for sampling, a number of methods have been proposed to improve generation efficiency.

Consistency Models~\cite{cm,ict,ect,geng2025improved,scm} first propose enforcing that every point along the trajectory maps to a consistent endpoint, enabling accelerated generation. 
ShortCut~\cite{shortcut} decomposes a single-step jump between two time steps into two sub-steps and supervises the single-step velocity using the average velocity of these two sub-steps. 
MeanFlow~\cite{mf} defines the marginal velocity learned in Flow Matching as the instantaneous velocity, and models the relationship between instantaneous velocity and average velocity to estimate the average velocity between two time steps, enabling direct mapping between them. 
Decoupled MeanFlow~\cite{lee2025decoupled} initializes the MeanFlow network using a pre-trained Flow Matching model and injects a second timestep into the final layer of the network. 
$\alpha$-Flow~\cite{zhang2025alphaflow} decomposes the MeanFlow training objective into two components: flow matching and consistency, and introduces a curriculum learning strategy to gradually transition the training objective from flow matching to consistency.

These methods center on enabling the model to directly learn mappings between different time steps along the same trajectory, a principle we define as the trajectory consistency objective. 
Our primary contribution is identifying and rectifying the detrimental effects of training-data-derived conditional velocities on this objective, a pervasive yet overlooked issue in existing frameworks.

\section{Preliminaries}

\textbf{Flow Matching.} 
Flow matching (FM) aims to learn a velocity field that transforms a prior distribution (e.g., Gaussian noise) to the data distribution. 
Let the data distribution be $p_{data}$, and the noise distribution be $p_{noise} \equiv \mathcal{N}(0, I)$. We define a standard linear interpolation schedule with $t \in[0,1]$ as the time index. 
Given a data sample $x \sim p_{data}$ and a noise sample $\epsilon \sim p_{noise}$, a flow path is defined by linear interpolation $x_t=(1-t)x + t\epsilon$. 
The velocity along this probability path is defined as $v_t = \epsilon - x$, also referred to as the conditional velocity $v_t = v(x_t \mid x, \epsilon)$. 
Since a given $x_t$ can be composed of various pairs of $x$ and $\epsilon$, multiple corresponding conditional velocities $v_t$ may exist at the same spatial-temporal point $x_t$. 
Flow Matching models the expectation of all possible conditional velocities at $x_t$ as the marginal velocity field $u_t=u(x_t, t) = \mathbb{E}[v_t \mid x_t]$, and employs a network $u_\theta$ parameterized by $\theta$ to approximate $u_t$. 
However, since computing $u_t$ is intractable, Flow Matching proposes that using $v_t$ as the training target is equivalent to fitting $u_t$, which can be formulated as:
\begin{equation}
    \mathbb{E}_{\substack{x, t, \epsilon}}
\left[\left\| u_\theta(x_t, t) - (\epsilon-x) \right\|_2^2\right].
\label{eq:eq1}
\end{equation}
The sampling is then performed by numerically solving the ODE governed by this marginal velocity field: $dx_t=u(x_t,t)dt$. 
This solution process can be viewed as integrating the flow backward in time. 
Formally, we define the flow map $\Phi_u$ to describe the trajectory; given two time steps $t$ and $s$,, the transition from state $x_t$ to $x_s$ can be written as:
\begin{equation}
     \Phi_u(x_t,s,t) = x_s = x_t - \int_{s}^{t} u(x_\tau, \tau)\, d\tau.
\label{eq:eq2}
\end{equation}
The standard flow matching sampling procedure typically starts from Gaussian noise $x_1 \sim \mathcal{N}(0,I)$ and gradually transitions to the data $x_0 \sim p_{data}$. 
This process is typically implemented using numerical solvers such as Euler~\cite{song2020denoising, scoresde} or Heun~\cite{karras2022elucidating} methods.
However, dozens or even hundreds of NEFs are usually required to ensure an accurate numerical approximation of the integral under a discretized time step.

\textbf{Fast Flow Models.} 
To further accelerate the sampling process, a fast flow model $F_\theta$ directly learns the average velocity between $x_t$ and $x_s$, thereby employing the linear mapping $f_\theta = x_t - (t-s)F_\theta$ to approximate the true marginal flow map $\Phi_u$. 
For example, MeanFlow interprets the marginal velocity as an instantaneous velocity and proposes approximating the integral of the velocity field $u$ divided by the time interval using the average velocity between time steps $s$ and $t$:
\begin{equation}
    u_{\text{avg}}(x_t,s,t) = \frac{1}{t-s}\int_{s}^{t} u(x_\tau, \tau)\, d\tau.
    \label{eq:eq3}
\end{equation}
By further deriving Eq.~\eqref{eq:eq3} and training a neural network $F_\theta$ to approximate the average velocity $u_{\text{avg}}$, we obtain the following expression:
\begin{equation}
    F_\theta(x_t,s,t)=u_t-(t-s) \frac{dF_\theta(x_t,s,t)}{dt},
    \label{eq:eq4}
\end{equation}
where the total derivative is defined as $\frac{d}{dt}F_\theta=\nabla x_t F_\theta\frac{dx_t}{dt} + \partial_t  F_\theta$, and $u_t=u(x_t,t)$ denotes the marginal velocity. 
Since the marginal velocity $u_t$ is not directly accessible during training, MeanFlow replaces it with the conditional velocity $v_t=\epsilon-x$. Consequently, the training objective is formulated as follows:
\begin{equation}
    \mathbb{E}_{x_t,v_t,s,t}\| F_\theta(x_t,s,t)-sg(v_t-(t-s)\dot{F}_\theta(v_t))\|^2,
    \label{eq:eq5}
\end{equation}
where $sg(\cdot)$ denotes the stop-gradient operation, and $\dot{F}_\theta(v_t)=\nabla x_t F_\theta \cdot v_t + \partial_t  F_\theta$ denote the total derivative of $F_\theta$ with respect to $t$ along the velocity field $v_t$. 
The intuition behind Eq.~\eqref{eq:eq5} is to enforce trajectory consistency. 
Ideally, if $F_\theta(x_t, s, t)$ accurately recovers the average velocity from $t$ to $s$, the predicted endpoint $f_\theta(x_t, s, t) = x_t - (t-s)F_\theta(x_t, s, t)$ should remain invariant to the starting time $t$. 
Consequently, minimizing Eq.~\eqref{eq:eq5} is equivalent to enforcing the consistency objective $\mathbb{E} \|\frac{d}{dt} f_\theta(x_t, s, t)\|^2 = \mathbb{E} \|\nabla_{x_t} f_\theta \cdot v_t + \partial_t f_\theta\|^2$.
Although the collection of all possible $v_t$ is equivalent to $u_t$ in expectation, we argue that using $v_t$ as instantaneous velocity estimates in fast flow models causes the training objective to deviate from the true marginal trajectory. 
As a result, the optimization target shifts from following a single marginal trajectory to approximating an expectation over multiple distinct trajectories, leading to suboptimal performance.

\begin{figure}[t]
    \centering
    \includegraphics[width=1.0\linewidth]{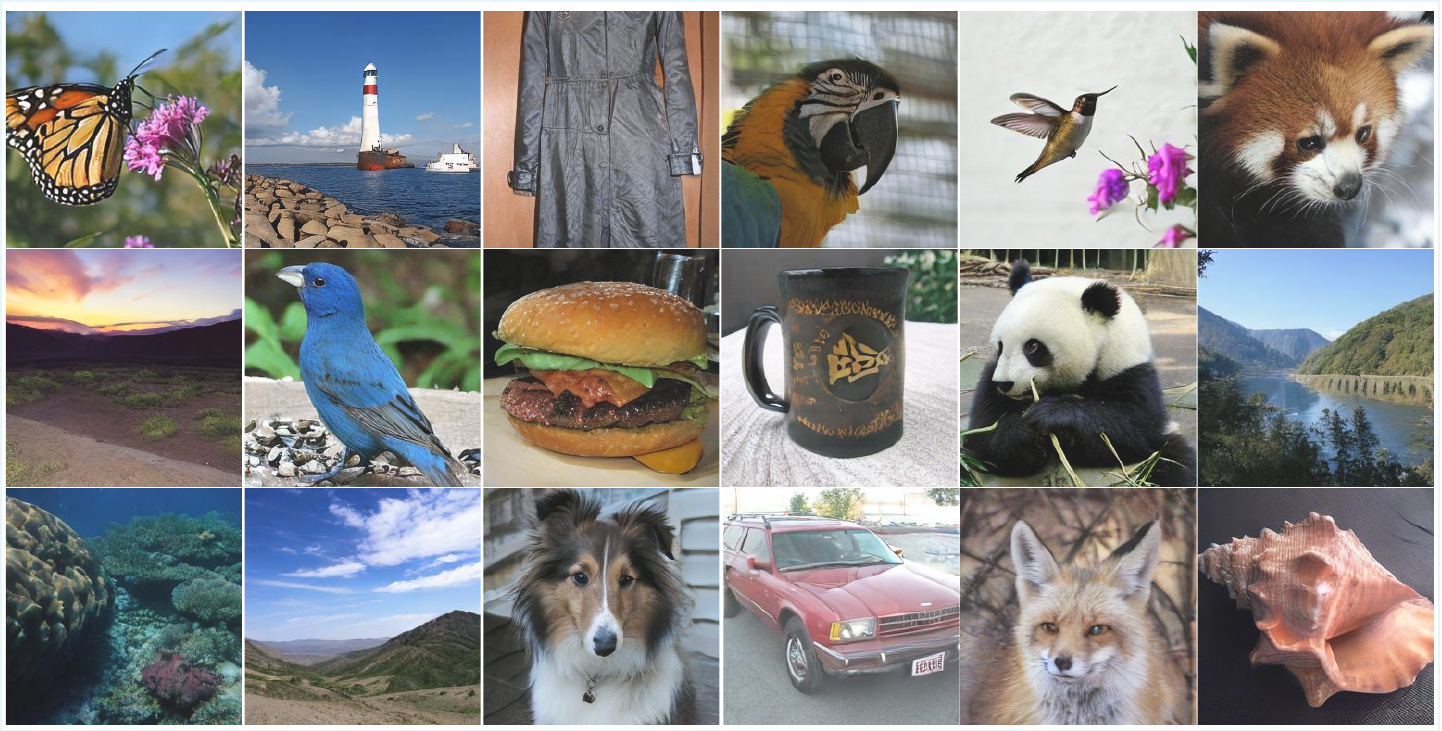}
    \caption{1-NFE generation results of FlowConsist-XL/2 on ImageNet 256x256. More uncurated results are in the Appendix~\ref{visual}.}
    \label{fig:visual}
    \vspace{-1em}
\end{figure}

\begin{figure*}[t]
    \centering
    \includegraphics[width=1.0\linewidth]{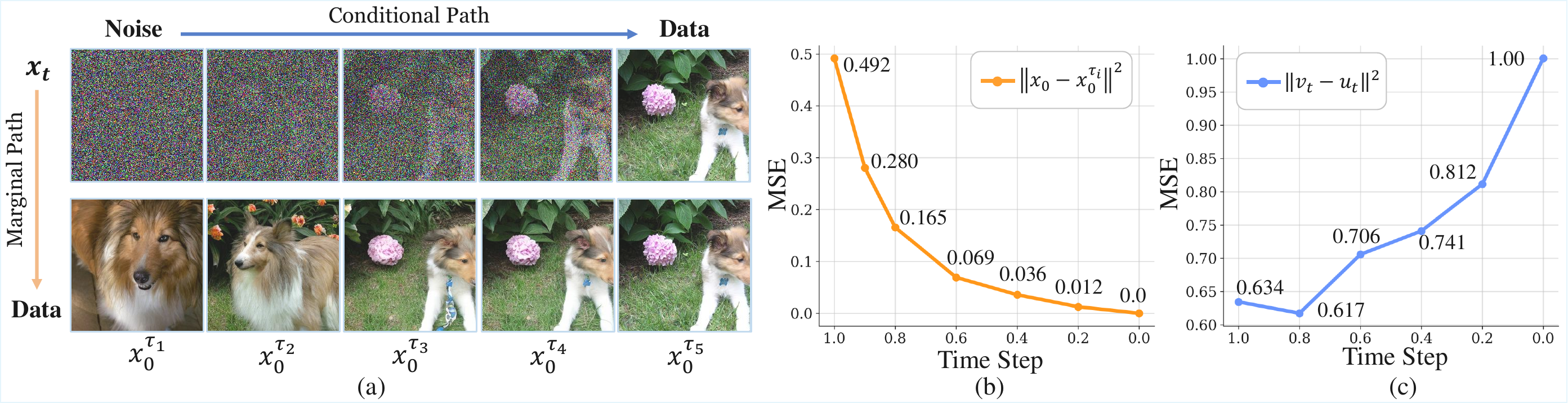}
    \caption{(a) Comparison between conditioned and marginal paths. We construct a conditioned path by adding random noise to an image $x_0$, and subsequently trace marginal paths $\tau_i$ starting from various time steps along this conditioned path using a pre-trained flow model. By comparing the data sample $x_0^{\tau_i}$ of these distinct trajectories, we observe that different points on a single conditioned path actually belong to divergent marginal paths $\tau$. (b) Mean Squared Error (MSE) between $x_0$ and different $x_0^{\tau_i}$, plotted as a function of the starting time step. The discrepancy between the data samples of the conditioned and marginal paths increases significantly as the time span grows larger. (c) The MSE between the marginal velocity $u_t$ and conditional velocity $v_t$ for various $x_t$ along the conditional path, plotted as a function of the time step. For any given $x_t$, there exists a persistent discrepancy between $u_t$ and $v_t$.}
    \label{fig:conditional path}
    \vspace{-1em}
\end{figure*}

\section{FlowConsist Model}
We identify two pervasive issues in current fast flow models.
(1) Theoretically, conditional velocities introduce multiple intersecting trajectories passing through $x_t$, which precludes a straight-line mapping between the start point $x_t$ and the end point $x_s$.
We provide a theoretical proof for this phenomenon and propose substituting the conditional velocity in the training objective with the model’s self-predicted marginal velocity (Sec.~\ref{sec:Flow Consistent}). 
(2) Practically, approximation errors induced by the model accumulate over the time span, leading to significant trajectory bias. 
To address this, we incorporate a marginal velocity learned directly from real data to rectify the training target (Sec.~\ref{sec:Flow Accurate}).

\subsection{Make Your Flow Consistent}
\label{sec:Flow Consistent}
Fast flow models achieve linear mapping from $x_t$ to $x_s$ by learning the average velocity between these time steps.
%
% We point out that such a linear mapping requires $x_t$ and $x_s$ to remain on a consistent trajectory, whereas Flow Matching fails to maintain this consistency, inevitably resulting in curved marginal paths. 
We point out that such a linear mapping requires $x_t$ and $x_s$ to remain on a consistent trajectory, whereas the flow matching model fails to maintain this consistency, inevitably resulting in curved marginal paths. 
% Flow Matching
The flow matching model is trained within a conditional velocity field where multiple pairs of $\epsilon$ and $x$ construct the conditional velocity $v_t$ for a given $x_t$. 
The sample $\hat{x}=x_t-t \cdot u_t$ obtained via linear mapping along the marginal velocity $u_t$ corresponds to the expectation of all possible $x$ given $x_t$ rather than a specific sample in the data distribution.
Given the uniqueness of solutions to ODE, marginal paths are uniquely determined and non-intersecting. 
Therefore, training a fast flow model in a marginal velocity field ensures that any $x_s$ corresponding to a given $x_t$ is uniquely defined, thereby enabling the linear mapping from $x_t$ to $x_s$.
Consequently, we emphasize that trajectory consistency is the core requirement for fast flow models to successfully execute linear mappings.

Denote the clean sample derived from a marginal path $\tau_i$ starting at $x_{t_i}$ as $x_0^{\tau_i} = \Phi_u(x_{t_i}, 0, t_i)$.
To illustrate the trajectory drift induced by conditional velocities within the marginal flow field, 
we visualize $x_0^{\tau_i}$ obtained by tracing marginal paths $\tau_i$ (predicted by a pre-trained model) from different samples $x_t$ along a single conditioned path constructed by linear interpolation between noise $\epsilon$ and image $x_0$, as shown in Fig.~\ref{fig:conditional path}(a).
As the noise level increases, $x_0^{\tau_i}$ exhibit increasingly significant deviations from $x_0$.
This further indicates that moving along a conditioned path entails crossing between marginal trajectories. 
Fig.~\ref{fig:conditional path}(b) quantifies the impact of this trajectory crossing by calculating the Mean Squared Error (MSE) between $x_0$ derived from conditioned paths and $x_0^{\tau_i}$. 
It illustrates that the further one moves along a conditioned path, the greater the impact of this trajectory drift. 
In fact, this phenomenon stems from the persistent discrepancy between conditional and marginal velocities at any given $x_t$. 
Accordingly, we calculate the MSE between the conditional and marginal velocity of $x_t$ as a function of $t$ in Fig.~\ref{fig:conditional path}(c). 
Notably, at $t=0$, the MSE between the conditional and marginal velocities is exactly 1, which corresponds to the MSE between Gaussian noise and its expectation $\mathbb{E}[\epsilon]=0$. 
This indicates that, at this point, the data sample can form conditional velocities with any arbitrary Gaussian noise, leading to the marginal velocity $u_0 = -x$. 
Therefore, this inconsistency does not stem from the flow model's fitting error of the marginal velocity.
In contrast, it reflects a theoretical and intrinsic difference between marginal and conditional velocities, which can be formulated as the following theorem.

\noindent \textbf{Theorem 1.} \textit{Let $x \sim p_{\text{data}}$ be a random variable supported on a data manifold $\mathcal{X} \subseteq \mathbb{R}^d$, and let $\epsilon \sim p_{\text{noise}}$ be a noise variable (e.g., Gaussian) independent of $x$. Consider the probability path $x_t = (1-t)x + t\epsilon$ for $t \in [0, 1]$. Let $v_t(x_t|x, \epsilon)$ denote the unique conditional velocity vector associated with a specific pair $(x, \epsilon)$, and let $u_t(x_t) = \mathbb{E}[v_t | x_t]$ denote the marginal velocity field at $x_t$.} \textit{Assuming the data distribution $p_{\text{data}}$ is non-degenerate (i.e., not a Dirac mass), the conditional covariance of the velocity $\Sigma_t(x_t) = \mathbb{E}[(v_t - u_t)(v_t - u_t)^\top \mid x_t]$ satisfies $\Sigma_t(x_t) \neq 0$ almost surely.}

The full proof of theorems is provided in Appendix~\ref{appendix}. 
Theorem 1 reveals the inherent inconsistency between $v_t$ and $u_t$ in Flow Matching.
For fast flow models, the inconsistency leads to a drift in the target trajectory.
To formalize how conditional training deviates from the marginal trajectory, let $F_\theta(x_t, s, t)$ be the average velocity predicted by the model from time $t$ to $s$ and $f_\theta(x_t, s, t) = x_t - (t-s)F_\theta(x_t, s, t)$ is the corresponding estimated flow mapping.
We further formulate this trajectory drift induced by conditional velocity in the following theorem.

\noindent \textbf{Theorem 2.} \textit{Let $v_t$ be the conditional velocity defined by the stochastic coupling $(x, \epsilon)$, and let $u_t$ be the corresponding marginal velocity field. Define the conditional covariance of the velocity as $\Sigma_t(x_t)$. In the continuous-time limit where $\Delta t \to 0$, let the conditional objective be defined as $\mathcal{L}_{cond}(\theta) = \mathbb{E}_{x_t, v_t} [ \| \nabla_{x_t} f_\theta \cdot v_t + \partial_t f_\theta \|^2 ]$. Then, the objective decomposes as:
% $$\mathcal{L}_{cond}(\theta) = \mathcal{L}_{consist}(\theta) + \mathcal{L}_{var}(\theta) $$
% $$\mathcal{L}_{consist}(\theta)=\mathbb{E}_{x_t} [ \| \nabla_{x_t} f_\theta \cdot u_t +\partial_t f_\theta \|^2 ] $$
% $$  \mathcal{L}_{var}(\theta)=\mathbb{E}_{x_t,v_t} [ \text{Tr}( \nabla_{x_t} f_\theta \Sigma_t(x_t) (\nabla_{x_t} f_\theta)^\top ) ].$$
\begin{equation*}
\begin{gathered}
\mathcal{L}_{\mathrm{cond}}(\theta)
= \mathcal{L}_{\mathrm{consist}}(\theta)
+ \mathcal{L}_{\mathrm{var}}(\theta), \\
\mathcal{L}_{\mathrm{consist}}(\theta)
= \mathbb{E}_{x_t}\!\left[
   \big\| \nabla_{x_t} f_\theta \cdot u_t
   + \partial_t f_\theta \big\|^2
   \right], \\
\mathcal{L}_{\mathrm{var}}(\theta)
= \mathbb{E}_{x_t,v_t}\!\left[
   \mathrm{Tr}\!\Big(
     \nabla_{x_t} f_\theta\,
     \Sigma_t(x_t)\,
     (\nabla_{x_t} f_\theta)^\top
   \Big)
   \right].
\end{gathered}
\end{equation*}
The optimization of $\mathcal{L}_{cond}(\theta)$ is equivalent to the optimization of trajectory consistency $\mathcal{L}_{consist}(\theta)$ if and only if $\Sigma_t(x_t) = 0$ for any non-degenerate $f_\theta$.}

As established by Theorem 1, $\Sigma_t(x_t) \neq \mathbf{0}$ holds almost surely, thereby confirming the inevitable presence of the trajectory perturbation effect denoted by $\mathcal{L}_{var}(\theta)$.
Theorem 2 explicitly identifies the issue arising from substituting marginal velocity with conditional velocity: the existence of $\mathcal{L}_{var}$ forces $\nabla_{x_t} f_\theta$ to vanish in high-variance directions. 
This gradient suppression prevents the model from faithfully fitting trajectories with significant curvature, resulting in systematic manifold drift. 
Consequently, we argue that a fast flow model consistent with the ODE trajectory can only be achieved by eliminating the influence of conditional velocity and optimizing solely based on $\mathcal{L}_{consist}$. 
Since $f_\theta(x_t,s,t) = x_t - (t-s)F_\theta(x_t,s,t)$, we derive the optimization objective $\mathcal{L}_{consist}$ for $F_\theta$ as:
\begin{equation}
    \mathbb{E}_{x_t} [ \| F_\theta(x_t,s,t) - u_t + (t-s)\dot{F}_\theta(u_t) \|^2 ],
    \label{eq:eq6}
\end{equation}
where $\dot{F}_\theta(u_t)=\nabla_{x_t} F_\theta \cdot u_t + \partial_t F_\theta$. Since $u_t$ is not directly accessible during training, an additional model would seem to be required for its estimation. 

However, we find that $u_t$ in the first term of Eq.~\eqref{eq:eq6} can be replaced by $v_t$, with the difference being only a constant term independent of the model parameters (we proof it in the Appendix~\ref{appendix}). 
In this case, the model naturally degenerates into the standard Flow Matching objective at $s=t$, which serves as a ready-made estimator for $u_t$ in the second term of Eq.~\eqref{eq:eq6}.
We thus arrive at the final training objective as follows:
\begin{equation}
\mathbb{E}_{x_t} [ \| F_\theta(x_t,s,t) - sg(v_t - (t-s)\dot{F}_\theta(u_\theta)) \|^2 ],    
\label{eq:eq7}
\end{equation}
where $v_t = \epsilon - x$, $u_\theta = F_\theta(x_t,t,t)$ and $sg(\cdot)$ is stop-gradient operation to avoid the computation of higher-order gradients as in previous works~\cite{scm,mf,tong2025flow,geng2025improved}. 
Eq.~\eqref{eq:eq7} consists of two components: the first is the standard Flow Matching term provided by $F_\theta - v_t$, and the second is the trajectory consistency regularization term $\nabla_{x_t} F_\theta \cdot u_\theta + \partial_t F_\theta$.
Although the conditional velocity is introduced, it does not induce trajectory drift during training. 
This is because once $F_\theta$ converges under the Flow Matching term at $s=t$, the term $\nabla_{x_t} F_\theta \cdot u_\theta + \partial_t F_\theta$ provides a stable correction for trajectory consistency.
The influence of $v_t$ on the training target eventually converges to its expectation $\mathbb{E}[v_t(x_t|x, \epsilon)]$, which is precisely the marginal velocity.

Additionally, we incorporate the Classifier-Free Guidance (CFG)~\cite{cfg} strategy into the model's training process. Following MeanFlow, we replace the original conditional velocity $v_t$ with $v_t^\omega = (\epsilon - x) + (1 - 1/\omega)(F_\theta(x_t, t, t | c) - F_\theta(x_t, t, t | \emptyset))$, where $c$ denotes the class condition, $\emptyset$ represents the null input, and $\omega$ is the effective CFG scale. Furthermore, we provide $\omega$ as a secondary input to the model, denoted as $F_\theta(x_t, t, t | c, \omega)$, enabling the model to achieve various CFG effects during inference simply by adjusting the value of $\omega$.

\subsection{Make Your Flow Accurate}
\label{sec:Flow Accurate}

While Eq.~\ref{eq:eq7} mitigates the trajectory drift caused by conditional velocity, the predicted trajectory $f_\theta$ remains susceptible to iterative approximation errors. 
In our consistency-based training, the optimization target at time $t$ is inherently coupled with the model's own predictions at the preceding time steps $t-\Delta t$. 
Consequently, any local bias in $f_\theta$ is not only retained but compounded through this self-bootstrapping mechanism. 
As $t$ increases, these infinitesimal errors propagate and accumulate, leading to a monotonic divergence from the ground-truth trajectory. 
We formalize this cumulative drift in the following theorem.

\begin{figure}
    \centering
    \includegraphics[width=1.0\linewidth]{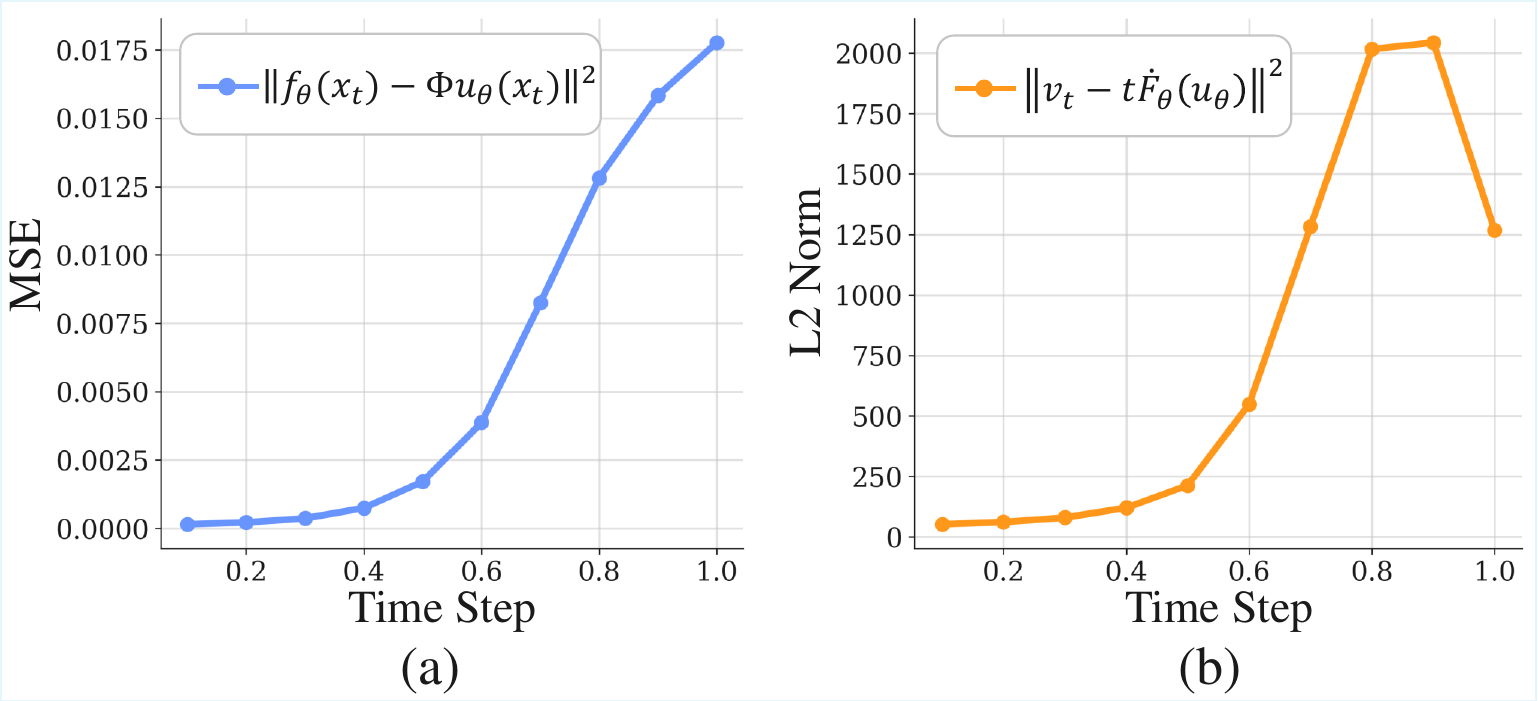}
    \caption{(a) MSE between the 1-NFE samples mapped from various time steps along the marginal trajectory and the corresponding ground truth data samples, plotted against the starting time step. This serves to quantify the cumulative error inherent in single-step mapping across different time steps. (b) During the actual training process, with $s=0$ fixed, the $L_2$ norm of the training objective defined by Eq.~\eqref{eq:eq7} as it varies with the starting time step $t$.}
    \label{fig:accumulate}
    \vspace{-1.5em}
\end{figure}

\noindent \textbf{Theorem 3.} \textit{Let $f^*(x_t, s,t)=x_s$ denote the ideal consistency mapping following the marginal trajectory $u_t$, and let $f_\theta(x_t,s,t)$ be the learned model. Let $e(s,t) = f_\theta(x_t, s, t) - f^*(x_t, s, t)$ denote the total approximation error from $s$ to $t$, $\mathbf{R}(t) = \frac{\partial f_\theta}{\partial t} + \nabla_{x_t} f_\theta \cdot u_t$ denote local approximation residual. The $e(s,t)$ evolves according to the following integral dynamics:}
$$e(s,t) = \int_s^t \mathbf{R}(r) dr.$$
To further demonstrate the impact of this error accumulation, we fix $s=0$ and compare the relative error between the trajectory directly predicted by the model $f_\theta(x_t, 0, t)$ and the trajectory obtained via sampling along the marginal velocity $f_\theta(x_t, t, t)$. 
As illustrated in Fig.~\ref{fig:accumulate}(a), the relative error grows monotonically with $t$, which is consistent with our theoretical analysis of cumulative error.
To characterize the impact of error accumulation during training, we plot the $L_2$ norm of the target velocity derived from Eq.~\eqref{eq:eq7} as a function of time steps in Fig.~\ref{fig:accumulate}(b). 
As the time step increases, the $L_2$ norm of the target velocity exhibits a pronounced upward trend, deviating significantly from the expected values. 
This observation suggests that the raw training objective fails to provide accurate guidance for the average velocity, which explains why adaptive weighting or normalization techniques are generally indispensable for ensuring training stability in fast flow models. 

Inspired by DMD~\cite{dmd}, we propose leveraging the marginal velocity learned from real data to rectify the mapping between two timesteps along the trajectory.
By aligning the marginal distribution of the model-generated data $p_{\text{fake}}$ with the real data distribution $p_{\text{real}}$, we ensure distributional consistency.
Specifically, we aim to minimize the KL divergence between $p_{\text{fake}}$ and $p_{\text{real}}$ across the entire diffusion process:
\begin{equation}
D_{\text{KL}}(p_{\text{fake}} \parallel p_{\text{real}})  = \mathbb{E}_{x_t, t} \left[ \log \frac{p{_\text{fake}}(x_t)}{p_{\text{real}}(x_t)} \right].
\label{eq:eq8}
\end{equation}
The optimization can be conceptualized as guiding the model using gradients derived from the discrepancy between the score functions of the real and fake distributions. 
Given the interconvertibility between the marginal velocity $u$ and the score function $\nabla \log p$, the gradient of the objective with respect to the model parameters $\theta$ can be formulated as:
\begin{equation}
\begin{aligned}
\nabla_\theta D_{\text{KL}} &= \mathbb{E}_{x_t,t}\left[ \left(\log p_{\text{fake}}(x_t) - \log p_{\text{real}}(x_t)\right) \frac{\partial x_t}{\partial \theta} \right] \\
&= \mathbb{E}_{x_t,t}\left[ w(t) \left( u_{\text{fake}}(x_t) - u_{\text{real}}(x_t) \right) \frac{\partial x_t}{\partial \theta} \right],
\end{aligned}
\label{eq:eq9}
\end{equation}
where $w(t)=\frac{t-1}{t}$ is a time-dependent weighting factor that facilitates the alignment of velocity fields.
We point out that this gradient can be interpreted as aligning the marginal velocity of model samples with the marginal velocity of real data. 
Since the marginal distribution of real data is generally inaccessible, DMD employs a pre-trained teacher model to estimate. 
We argue that it suffices to learn the marginal velocity induced by real data. 
Therefore, we replace the teacher-estimated marginal velocity with the marginal velocity predicted by the model itself.
While DMD focuses on learning a direct mapping from noise to data, our approach requires learning mappings from arbitrary time steps along the marginal trajectory.
Consequently, we apply the marginal velocity alignment to the mappings from any given time step $t$ to the data, thereby rectifying the cumulative effect of model approximation errors along the trajectory. 
Although we can substitute the pre-trained teacher model with the model predicted marginal velocity, it remains necessary to train an auxiliary model $G_\psi$ to estimate the marginal velocity of the samples predicted by the model. 
Furthermore, since these samples may originate from different time steps $t$ on the trajectory and incorporate noise from another time step $t'$, we denote such samples as $x_{t'}^t = t' \epsilon + (1-t')  x_0^t$, where $x_0^t = f_\theta(x_t, 0, t)$. 
Similar to Flow Matching, we train $G_\psi$ using conditional velocity to estimate the marginal velocity for each sample along the trajectory.
\begin{equation}
    \mathbb{E}_{\substack{x^t_0,t, t', \epsilon}}
\left[\left\| G_\psi(x_{t'}^{t}, t, t') - (\epsilon-x_0^t) \right\|_2^2\right].
\label{eq:eq10}
\end{equation}
Subsequently, we leverage the marginal velocity $F_\theta$, learned from ground-truth data, to align with the marginal velocity $G_\psi$ estimated from model samples. 
This process aims to correct the cumulative approximation errors encountered when the model performs single-step mappings to the data along the trajectory, and can be formulated as follows:
\begin{equation}
   \mathbb{E}_{x_t,t,t',\epsilon} [ \| F_\theta(x_t,0,t) - sg(F_\theta(x_{t'}^t,t',t')-G_\psi(x_{t'}^t,t,t')) \|^2 ].
    \label{eq:eq11}
\end{equation}
Since $F_\theta(x_t, 0, t)$ is continuously optimized by the error-prone target velocity in Eq.~\eqref{eq:eq7} without direct exposure to the real data distribution, Eq.~\eqref{eq:eq11} can be interpreted as pulling it directly toward the real marginal distribution via $F_\theta(x_t, t', t')$, thereby correcting the accumulated error.
Notably, although Eq.~\eqref{eq:eq11} does not explicitly optimize the global average velocity $F_\theta(x_t, 0, 1)$, this correction propagates to $F_\theta(x_t, 0, 1)$ through Eq.~\eqref{eq:eq5}, ultimately enhancing model performance.

\section{Experiment}
We conducted experiments on the ImageNet dataset at $256 \times 256$ resolution to validate the proposed method. 
Following previous approaches~\cite{mf,lipman2023flow,cm,scm,geng2025improved}, we perform image generation in the latent space of a pre-trained VAE encoder~\cite{rombach2022high} and utilize the DiT~\cite{dit} architecture. 
For ablation studies, we initialize our model $F_\theta$ and $G_\psi$ with a pre-trained SiT-B/2~\cite{ma2024sit} and train it for 300K iterations with a batch size of 256. 
We set the CFG scale $\omega$ by sampling uniformly from the interval $[1, 4]$.
The model's performance is evaluated using the 1-NFE generation setting, with the Fréchet Inception Distance (FID)~\cite{fid} measured on 50K generated samples. 
Detailed experimental configurations are provided in the Appendix~\ref{details}.

\begin{table*}[t]
\centering
\caption{\textbf{Ablation studies on 1-NFE generation.} FID is evaluated on ImageNet $256\times256$ using 50k samples.}
% \vspace{8pt}

\begin{minipage}[b]{0.32\linewidth}
    \centering
    \footnotesize
    \setlength{\tabcolsep}{3pt}
    \begin{tabular}{y{90pt}|x{30pt}}
    \textbf{Methods} & FID \\
    \hline
    original MF & 6.17 \\
    with CFG-condition & 5.28 \\
    with Eq.~\eqref{eq:eq7} & 4.64 \\
    with Eq.~\eqref{eq:eq7} \& \eqref{eq:eq11} & \textbf{4.31} \\
    \end{tabular}
    \vspace{2pt}
    \subcaption{MeanFlow Framework}
    \label{tab:ablation_mf}
\end{minipage}
\hfill
\begin{minipage}[b]{0.33\linewidth}
    \centering
    \footnotesize
    \setlength{\tabcolsep}{3pt}
    \begin{tabular}{y{70pt}|x{30pt}}
    \textbf{Methods} & FID \\
    \hline
    10\% ratio & 4.58 \\
    20\% ratio & 4.48 \\
    30\% ratio & \textbf{4.31} \\
    40\% ratio & 4.41 \\
    \end{tabular}
    \vspace{2pt}
    \subcaption{Integration Ratio of Eq.~\eqref{eq:eq7} and Eq.~\eqref{eq:eq11}}
    \label{tab:ablation_eq11}
\end{minipage}
\hfill
\begin{minipage}[b]{0.30\linewidth}
    \centering
    \footnotesize
    \setlength{\tabcolsep}{3pt}
    \renewcommand{\arraystretch}{1.25} 
    \begin{tabular}{y{100pt}|x{30pt}}
    \textbf{Methods} & FID \\
    \hline
    CM with CFG condition & 6.74 \\
    with Eq.~\eqref{eq:eq7} & 5.84 \\
    with Eq.~\eqref{eq:eq7} \& \eqref{eq:eq11} & \textbf{5.36} \\
    \end{tabular}
    \vspace{2pt}
    \subcaption{Consistency Model}
    \label{tab:ablation_scm}
\end{minipage}
\label{tab:ablation}
\vspace{-1em}
\end{table*}

\subsection{Ablation Study}
\begin{figure}
    \centering
    \includegraphics[width=1.0\linewidth]{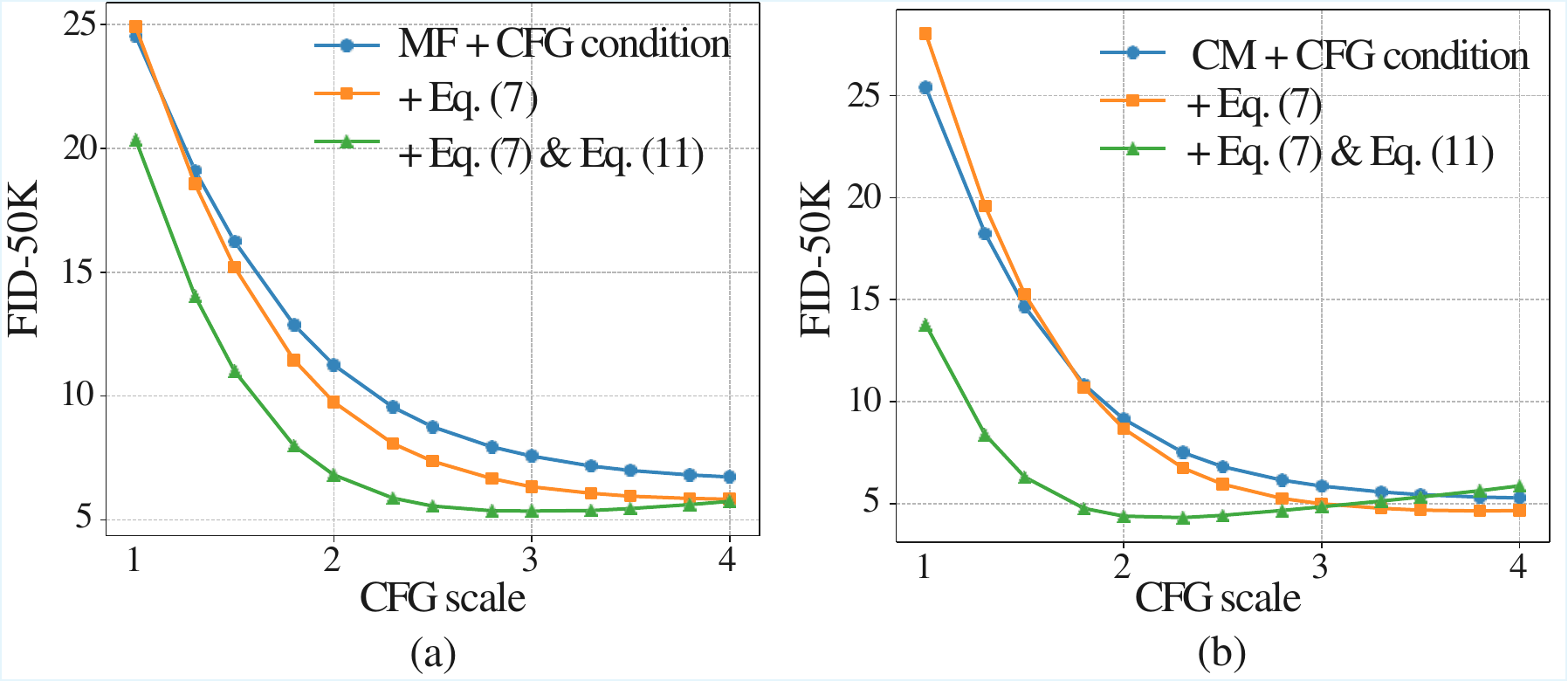}
    \caption{FID curves as a function of CFG scale under different configurations. (a) Using MeanFlow (MF) as the model architecture: FID scores versus CFG scale after sequentially applying CFG conditioning (best FID: 5.28), trajectory consistency loss in Eq.~\ref{eq:eq7} (best FID: 4.64), and marginal velocity alignment loss in Eq.~\ref{eq:eq11} (best FID: 4.31). (b) Using Consistency Models (CM) as the architecture: the best FID scores for CFG conditioning, trajectory consistency loss, and marginal velocity alignment loss are 6.74, 5.84, and 5.36, respectively.}
    \label{fig:cfg-fid}
    \vspace{-1.5em}
\end{figure}

\noindent \textbf{CFG condition.} We incorporate the CFG scale $\omega$ as a conditional input to the network, encoded in the same format as the time steps $t$. 
In Tab.~\ref{tab:ablation}(a), we validate the effectiveness of this CFG conditioning, where the reported FID scores are the optimal values selected from the CFG-FID curves shown in Fig.~\ref{fig:cfg-fid}. 
By integrating CFG conditioning, we not only gain the ability to controllably adjust the CFG scale during inference but also observe a significant improvement in overall performance compared to the original MeanFlow~\cite{mf}. 
This improvement likely stems from the broader range of CFG values, which bolsters the model's generalization and facilitates the search for the optimal FID. 
In subsequent experiments, we maintain this CFG conditioning and consistently report the best FID across all scales.

\noindent \textbf{Marginal trajectory consistency loss in Eq.~\eqref{eq:eq7}.} 
We replace the average velocity loss in MeanFlow~\cite{mf} with our proposed marginal trajectory consistency loss described in Sec~\ref{sec:Flow Consistent}.
Tab.~\ref{tab:ablation}(a) shows that the FID improves from 5.86 to 4.99, a solid gain of 0.87.
Furthermore, Fig.~\ref{fig:cfg-fid} illustrates the FID variance across different CFG scales; our method outperforms the mean velocity loss in MeanFlow as the CFG scale increases, maintaining a consistent lead at larger CFG scales.
This superiority arises from our model’s ability to avoid direction drift induced by conditional velocities, ensuring that optimization remains aligned with a consistent marginal trajectory target.

\noindent \textbf{Trajectory marginal velocity alignment in Eq.~\eqref{eq:eq11}.}
As shown in Tab.~\ref{tab:ablation}(a), we conducted ablation studies on the marginal velocity alignment loss proposed in Sec.~\ref{sec:Flow Accurate}. 
By employing samples generated with the marginal velocities aligned to real data, the FID further reduces from 4.99 to 4.34. 
Notably, upon applying marginal velocity alignment, the model's FID decreases and then significantly increases with the CFG scale, whereas other models exhibit a trend of decreasing FID followed by a plateau.
This suggests that a higher CFG scale may help provide a clearer direction for the marginal path, thereby partially mitigating the error accumulation in Eq.~\eqref{eq:eq7}.
However, once the accumulated error is corrected by Eq.~\eqref{eq:eq11}, an excessively high CFG can lead to a reduction in diversity and a subsequent increase in FID, much like the behavior of multi-step Diffusion~\cite{ncsn,scoresde,karras2022elucidating} or Flow-based models~\cite{rectified,xu2022poisson,lipman2023flow,xu2023pfgm++}.

\begin{table*}[!ht]
    \centering
    \caption{
        \textbf{Comparison on class-conditional ImageNet 256$\bm{\times}$256.} \textbf{Left:} Fast flow models trained distilled from a pre-trained teacher and without distillation. \textbf{Right:} Diffusion/flow models with multi NFE. All results incorporate the CFG strategy, $\times2$ denotes that the use of CFG doubles the NFE during inference.
    }
    \vspace{-0.8em}
    \label{tab:main}
    \scriptsize
    \setlength{\tabcolsep}{8pt}

    \begin{subtable}[t]{0.48\linewidth}
    \centering
    \begin{NiceTabular}[t]{
        l
        c
        c
        c
        S[table-format=2.2, detect-weight=true, mode=text]
    }
        \multicolumn{4}{l}{} \\
        \toprule
        \textbf{Method} & \textbf{NFE} $\downarrow$  &  \textbf{Epochs}  & \multicolumn{1}{c}{\textbf{FID} $\downarrow$}\\
        \midrule
        
        \multicolumn{5}{l}{\textit{\textbf{Fast Flow Model by distillation}}} \\
        \arrayrulecolor{black!30}\midrule
        SDEI~\cite{stei} & 8 & 20 & 2.46 \\
        \arrayrulecolor{black!10}\midrule
        FACM~\cite{facm} & 2 & --  & 1.52 \\
        \arrayrulecolor{black!10}\midrule
        \multirow{2}{*}{$\pi$-Flow~\cite{pi-flow}} & 1 & 448 & 2.85 \\
         & 2 & 448 & 1.97 \\
        \arrayrulecolor{black!10}\midrule
        FreeFlow-XL/2~\cite{tong2025flow} & 1 &  300  & \bfseries  1.45 \\
         
        \arrayrulecolor{black}\midrule
        
        \multicolumn{5}{l}{\textit{\textbf{Fast Flow Model without distillation}}} \\
        \arrayrulecolor{black!30}\midrule
        \multirow{2}{*}{Shortcut-XL/2~\cite{shortcut}} & 1 & 250 & 10.60\\
        & 128 & 250 & 3.80\\
        \arrayrulecolor{black!10}\midrule
        \multirow{2}{*}{IMM-XL/2~\cite{imm}} & 1${\times}$2 & 3840 & 7.77\\
         & 8${\times}$2 & 3840 & 1.99\\
        \arrayrulecolor{black!10}\midrule
        \multirow{2}{*}{STEI~\cite{stei}} & 1 & 1420 & 7.12\\
         & 8 & 1420 & 1.96\\
        \arrayrulecolor{black!10}\midrule
        \multirow{2}{*}{MeanFlow-XL/2~\cite{mf}} & 1 &  240 & 3.43\\
         & 2 & 1000 & 2.20 \\
        \arrayrulecolor{black!10}\midrule
        TiM-XL/2~\cite{tim} & 1 & 300 & 3.26\\
        \arrayrulecolor{black!10}\midrule
        DMF-XL/2~\cite{lee2025decoupled} & 1 & 880 & 2.16\\
        \arrayrulecolor{black!10}\midrule
        $\alpha$-Flow-XL/2~\cite{zhang2025alphaflow} & 1 & 300 & 2.58\\
        \arrayrulecolor{black!10}\midrule
        iMF-XL/2~\cite{geng2025improved} & 1 & 800 & 1.72\\
        \arrayrulecolor{black!10}\midrule
        FlowConsist-XL/2 & 1 & 200 & \bfseries 1.52\\
        \arrayrulecolor{black}\bottomrule

    \end{NiceTabular}
    \end{subtable}
    \hfill
    \begin{subtable}[t]{0.48\linewidth}
    \centering
        \begin{NiceTabular}[t]{
        l
        c
        c
        c
        S[table-format=2.2, detect-weight=true, mode=text]
    }
        \multicolumn{4}{l}{} \\
        \toprule
        \textbf{Method} & \textbf{NFE} $\downarrow$ & \textbf{Epochs} & \multicolumn{1}{c}{\textbf{FID} $\downarrow$} \\
        \midrule
        \multicolumn{5}{l}{\textit{\textbf{Multi-Step diffusion/flow Model}}} \\
        \arrayrulecolor{black!30}\midrule
        ADM-G~\cite{dhariwal2021diffusion} & 250${\times}$2 & 396 & 4.59 \\
        \arrayrulecolor{black!10}\midrule
        SimDiff~\cite{hoogeboom2023simple} & 1000${\times}$2 & - & 2.77 \\
        \arrayrulecolor{black!10}\midrule
        U-ViT-H/2~\cite{ma2024learning} & 50${\times}$2 & 400 & 2.29 \\
        \arrayrulecolor{black!10}\midrule
        DiT-XL/2~\cite{dit} & 250${\times}$2 & 1400 & 2.27 \\
        \arrayrulecolor{black!10}\midrule
        SiT-XL/2~\cite{ma2024sit} & 250${\times}$2 & 1400 & 2.06 \\
        \arrayrulecolor{black!10}\midrule
        MDT~\cite{mdt} & 250${\times}$2 & 1300 & 1.79 \\
        \arrayrulecolor{black!10}\midrule
        SiD2~\cite{hoogeboom2024simpler} & 512${\times}$2 & - & 1.38 \\
        \arrayrulecolor{black!10}\midrule
        \multirow{2}{*}{SiT-XL/2+REPA~\cite{repa}} & 250${\times}$2 & 200 & 1.96  \\
         & 250${\times}$2 & 800 & 1.42 \\
        \arrayrulecolor{black!10}\midrule
        \multirow{2}{*}{Light.DiT~\cite{yao2025reconstruction}} & 250${\times}$2 & 64 & 2.11  \\
         & 250${\times}$2 & 800 & 1.35 \\
        \arrayrulecolor{black!10}\midrule
        \multirow{2}{*}{DDT-XL~\cite{wang2025ddt}} & 250${\times}$2 & 80 & 1.52  \\
         & 250${\times}$2 & 400 & 1.26 \\
         \arrayrulecolor{black!10}\midrule
         $\text{DiT}^\text{{DH}}$-XL~\cite{zheng2025diffusion} & 250${\times}$2 & 1300 & 1.13 \\
        \arrayrulecolor{black}\midrule
         
        \multicolumn{5}{l}{\textit{\textbf{GANs}}} \\
        \arrayrulecolor{black!30}\midrule
        BigGAN~\cite{brock2018large} & 1 & - & 6.95 \\
        \arrayrulecolor{black!10}\midrule
        GigaGAN~\cite{kang2023scaling} & 1 & - & 3.45 \\
        \arrayrulecolor{black!10}\midrule
        StyleGAN-XL~\cite{sauer2022stylegan} & 1 & - & 2.30 \\
 
        \arrayrulecolor{black}\bottomrule
    \end{NiceTabular}
    \end{subtable}
    \vspace{-1em}
\end{table*}

\noindent \textbf{Integration of Eq.~\eqref{eq:eq7} and Eq.~\eqref{eq:eq11}}
Following previous work~\cite{shortcut, mf, tong2025flow}, we divide each training batch into two segments to independently execute the trajectory consistency objective and the marginal velocity alignment objective. 
By varying the ratio of these segments within the batch, we can control the respective contributions of these two losses to the gradient signal. 
As demonstrated in Tab.~\ref{tab:ablation}(b), the model's FID exhibits a trend of initially decreasing and then increasing, while showing overall robustness to the specific ratio of the two components. 
This indicates that even when the marginal velocity alignment loss accounts for a relatively small proportion of the training batch, its inclusion still yields a significant improvement in performance.

\noindent \textbf{Ablation on continuous-time consistency model.}
We apply our method to the training of Continuous Consistency Models and conduct ablation studies. 
By setting $s=0$ in the MeanFlow~\cite{mf} framework, any sample $x_t$ on the marginal trajectory can be directly mapped to the same clean data sample $x_0$ via $f_\theta(x_t,0,t) = x_t - t \cdot F_\theta(x_t,0,t)$.
Meanwhile, to ensure the model can predict the marginal velocity at $x_t$, we set $s=t$ with 60\% probability, enabling the model to provide marginal velocity estimates via $F_\theta(x_t,t,t)$. 
As shown in Tab.~\ref{tab:ablation}(c), our proposed strategy remains effective under the CM training paradigm. 
Incorporating the marginal trajectory consistency objective improves the FID from 6.74 to 5.84, while the addition of the marginal velocity alignment strategy further enhances the FID to 5.36. 
We also plot the FID scores against varying CFG scales in Fig.~\ref{fig:cfg-fid}(b), which exhibits similar trends to those observed in MeanFlow. 
These results demonstrate that our strategy is compatible with other fast-flow architectures. 
By mitigating trajectory drift and correcting error accumulation across time steps, existing fast-flow models can benefit from this training suite.

\subsection{Comparison with Other Methods} 

In Tab.~\ref{tab:main}, we compare our method with existing fast-flow models~\cite{geng2025improved,mf,tong2025flow,shortcut,lee2025decoupled,zhang2025alphaflow,tim} on the class-conditional ImageNet $256 \times 256$ dataset.
The second-best FID among prior methods is 1.72, achieved by iMF-XL/2~\cite{geng2025improved}, whereas our method reaches a state-of-the-art FID of 1.52, reducing the FID by 0.20.
This significant performance leap stems from our ability to rectify two types of biases affecting trajectory consistency. 
First, our method addresses objective trajectory deviation induced by conditional velocity, which theoretically constrains the performance upper bound of fast-flow models without distillation by shifting the optimal solution toward the average of multiple divergent trajectories.
Second, we mitigate the cumulative approximation error; the accumulation of residuals over large time spans causes substantial deviations in long-range direct mappings, thereby degrading practical performance. By correcting both biases. 
Our method ensures that the model consistently optimizes along the correct and unified trajectory, ultimately achieving superior performance. 
Compared to distillation-based methods, our approach demonstrates comparable results, effectively bridging the gap between the distillation and teacher-free paradigms in fast-flow modeling.

\section{Conclusion}
In this work, we identify the trajectory drift introduced by conditional velocities in fast-flow models, which prevents the optimization from aligning with a consistent marginal trajectory. 
To address this, we replace the conditional velocity in the partial differentiation with the marginal velocity predicted by the model itself, ensuring that the optimization objective remains anchored to a consistent marginal path.
Furthermore, we analyze the error accumulation during training and introduce a marginal velocity alignment strategy. 
This strategy aligns the clean samples predicted by the model with real data distributions by leveraging marginal velocities learned from real data. 
Experimental results demonstrate the effectiveness of our approach and its broad applicability to existing fast-flow frameworks.

\nocite{langley00}

\bibliography{references}
\bibliographystyle{icml2026}

%%%%%%%%%%%%%%%%%%%%%%%%%%%%%%%%%%%%%%%%%%%%%%%%%%%%%%%%%%%%%%%%%%%%%%%%%%%%%%%
%%%%%%%%%%%%%%%%%%%%%%%%%%%%%%%%%%%%%%%%%%%%%%%%%%%%%%%%%%%%%%%%%%%%%%%%%%%%%%%
% APPENDIX
%%%%%%%%%%%%%%%%%%%%%%%%%%%%%%%%%%%%%%%%%%%%%%%%%%%%%%%%%%%%%%%%%%%%%%%%%%%%%%%
%%%%%%%%%%%%%%%%%%%%%%%%%%%%%%%%%%%%%%%%%%%%%%%%%%%%%%%%%%%%%%%%%%%%%%%%%%%%%%%
\newpage
\appendix
\onecolumn
\section{Theoretical Analysis and Derivations.}\label{appendix}

\subsection{Theorem Proofs.}
\noindent \textbf{Theorem 1.} \textit{Let $x \sim p_{\text{data}}$ be a random variable supported on a data manifold $\mathcal{X} \subseteq \mathbb{R}^d$, and let $\epsilon \sim p_{\text{noise}}$ be a noise variable (e.g., Gaussian) independent of $x$. Consider the probability path $x_t = (1-t)x + t\epsilon$ for $t \in [0, 1]$. Let $v_t(x_t|x, \epsilon)$ denote the unique conditional velocity vector connecting a specific pair $(x, \epsilon)$, and let $u_t(x_t) = \mathbb{E}[v_t | x_t]$ denote the marginal velocity field at $x_t$.}\textit{Assuming the data distribution $p_{\text{data}}$ is non-degenerate (i.e., not a Dirac mass), the conditional covariance of the velocity $\Sigma_t(x_t) = \mathbb{E}[(v_t - u_t)(v_t - u_t)^\top \mid x_t]$ satisfies $\Sigma_t(x_t) \neq 0$ almost surely.}

\textit{Proof.} We analyze the boundary case $t=0$ and the interval $t \in (0, 1]$ separately.

\noindent \textbf{Case 1: $t=0$.} At $t=0$, the coupling $x_t = (1-t)x + t\epsilon$ collapses to $x_0 = x$. In this case, the data point is uniquely determined by the observation, giving $p(x \mid x_0) = \delta(x - x_0)$. The conditional velocity is $v_0 = \epsilon - x_0$. Since $\epsilon \sim \mathcal{N}(0, \mathbf{I})$ is independent of $x_0$, the conditional covariance is exactly the noise prior:
\begin{equation}
    \Sigma_0(x_0) = \mathbb{E}[(v_t - \mathbb{E}[v_t])(v_t - \mathbb{E}[v_t])^\top \mid x_t] = \text{Var}(v_0 \mid x_0) = \text{Var}(\epsilon) = \mathbf{I}
\end{equation}
\noindent \textbf{Case 2: $t \in (0, 1]$.} For $t \in (0, 1]$, substituting $\epsilon = \frac{x_t - (1-t)x}{t}$ into $v_t = \epsilon - x$, we obtain the linear relation $v_t = \frac{1}{t}(x_t - x)$.
Consequently, the conditional covariance of the velocity is proportional to the posterior variance of the data point:
\begin{equation}\Sigma_t(x_t) = \text{Var}(v_t \mid x_t) = \frac{1}{t^2} \text{Var}(x \mid x_t)\end{equation}
The posterior distribution satisfies $p(x \mid x_t) \propto p_{\text{data}}(x) \cdot \mathcal{N}\left( \frac{x_t - (1-t)x}{t}; 0, \mathbf{I} \right)$. Because $p_{\text{noise}}$ has full support on $\mathbb{R}^d$ and $p_{\text{data}}$ is non-degenerate (i.e., its support contains at least two distinct points), the posterior $p(x \mid x_t)$ cannot be a Dirac measure for almost all $x_t$. Therefore:
\begin{equation}
    \text{Var}(x \mid x_t) \neq \mathbf{0} \implies \text{Tr}(\Sigma_t(x_t)) > 0 \implies \Sigma_t(x_t) \neq 0\quad \text{a.s.}
\end{equation}
This confirms that $\Sigma_t(x_t) \neq 0$ throughout the interval $[0, 1]$.

\noindent \textbf{Theorem 2.} \textit{Let $v_t=v_t(x_t|x,\epsilon)$ be the conditional velocity defined by the stochastic coupling $(x, \epsilon)$, and let $u_t = \mathbb{E}[v_t | x_t]$ be the corresponding marginal velocity field. Define the conditional covariance of the velocity as $\Sigma_t(x) = \mathbb{E}[(v_t - u_t)(v_t - u_t)^\top | x_t]$. In the continuous-time limit where $\Delta t \to 0$, let the conditional objective be defined as $\mathcal{L}_{cond}(\theta) = \mathbb{E}_{x_t, v_t} [ \| \nabla_{x_t} f_\theta \cdot v_t + \partial_t f_\theta \|^2 ]$. Then, the objective decomposes as:$$\mathcal{L}_{cond}(\theta) = \mathcal{L}_{consist}(\theta) + \mathcal{L}_{var}(\theta) $$
$$\mathcal{L}_{consist}(\theta)=\mathbb{E}_{x_t} [ \| \nabla_{x_t} f_\theta \cdot u_t +\partial_t f_\theta \|^2 ] $$
$$  \mathcal{L}_{var}(\theta)=\mathbb{E}_{x_t,v_t} [ \text{Tr}( \nabla_{x_t} f_\theta \Sigma_t(x_t) (\nabla_{x_t} f_\theta)^\top ) ].$$The optimization of $\mathcal{L}_{cond}(\theta)$ is equivalent to the optimization of trajectory consistency $\mathcal{L}_{consist}(\theta)$ if and only if $\Sigma_t(x_t) = 0$ for any non-degenerate $f_\theta$.}

\textit{Proof.} To simplify the notation, let $\mathbf{J}_\theta(x_t, t) = \nabla_{x_t} f_\theta(x_t, t)$ denote the Jacobian matrix and $\dot{f}_\theta = \partial_t f_\theta(x_t, t)$ denote the partial time derivative. The conditional objective can be written as:
\begin{equation}
\mathcal{L}_{cond}(\theta) = \mathbb{E}_{x_t} \left[ \mathbb{E}_{v_t | x_t} [ || \mathbf{J}_\theta v_t + \dot{f}_\theta ||^2 ] \right]
\end{equation}

By adding and subtracting the term $\mathbf{J}_\theta u_t$, where $u_t = \mathbb{E}[v_t | x_t]$, the inner term becomes:
\begin{equation}
\begin{aligned}
  ||\mathbf{J}_\theta v_t + \dot{f}_\theta ||^2 &= || (\mathbf{J}_\theta u_t + \dot{f}_\theta) + \mathbf{J}_\theta (v_t - u_t) ||^2  \\
  & = ||\mathbf{J}_\theta u_t + \dot{f}_\theta ||^2 + 2(\mathbf{J}_\theta u_t + \dot{f}_\theta)^\top \mathbf{J}_\theta (v_t - u_t)  + || \mathbf{J}_\theta (v_t - u_t) ||^2    
\end{aligned}
\end{equation}

Now, we take the conditional expectation $\mathbb{E}_{v_t | x_t} [\cdot]$. The cross-product term vanishes because:
\begin{equation}
\mathbb{E}_{v_t | x_t} [\mathbf{J}_\theta (v_t - u_t)] = \mathbf{J}_\theta (\mathbb{E}[v_t | x_t] - u_t) = \mathbf{J}_\theta (u_t - u_t) = 0
\end{equation}
The remaining terms are:
\begin{equation}
\mathbb{E}_{v_t | x_t} [ || \mathbf{J}_\theta v_t + \dot{f}_\theta ||^2 ] = || \mathbf{J}_\theta u_t + \dot{f}_\theta ||^2 + \mathbb{E}_{v_t | x_t} [ || \mathbf{J}_\theta (v_t - u_t) ||^2 ]
\end{equation}
The second term can be rewritten using the property $\| z \|^2 = \text{Tr}(z z^\top)$:
\begin{equation}
\begin{aligned}
\mathbb{E}_{v_t | x_t} [ || \mathbf{J}_\theta (v_t - u_t) ||^2 ] & = \mathbb{E}_{v_t | x_t} [ \text{Tr}( \mathbf{J}_\theta (v_t - u_t) (v_t - u_t)^\top \mathbf{J}_\theta^\top ) ] \\
& = \text{Tr}( \mathbf{J}_\theta \mathbb{E}_{v_t | x_t} [ (v_t - u_t) (v_t - u_t)^\top ] \mathbf{J}_\theta^\top ) \\
& = \text{Tr}( \mathbf{J}_\theta \Sigma_t(x_t) \mathbf{J}_\theta^\top )
\end{aligned}
\end{equation}
Integrating over $p(x_t)$, we obtain the stated decomposition:
\begin{equation}
\mathcal{L}_{cond}(\theta) = \underbrace{\mathbb{E}_{x_t} [ || \nabla_{x_t} f_\theta \cdot u_t + \partial_t f_\theta ||^2 ]}_{\mathcal{L}_{consist}(\theta)} + \underbrace{\mathbb{E}_{x_t,v_t} [ \text{Tr}( \nabla_{x_t} f_\theta \Sigma_t(x_t) (\nabla_{x_t} f_\theta)^\top ) ]}_{\mathcal{L}_{var}(\theta)}
\end{equation}
Finally, $\mathcal{L}_{var}(\theta)$ is a quadratic form with respect to the Jacobian. The equality $\mathcal{L}_{cond}(\theta) = \mathcal{L}_{consist}(\theta)$ holds if and only if $\Sigma_t(x_t) = 0$ for any non-degenerate $f_\theta$.

\noindent \textbf{Theorem 3.} \textit{Let $f^*(x_t, s,t)=x_s$ be the ideal consistency mapping following the marginal trajectory $u_t$, and $f_\theta(x_t,s,t)$ be the learned model. Let $e(s,t) = f_\theta(x_t, s, t) - f^*(x_t, s, t)$ denote the total approximation error from $s$ to $t$, $\mathbf{R}(t) = \frac{\partial f_\theta}{\partial t} + \nabla_{x_t} f_\theta \cdot u_t$ denote local approximation residual. The $\epsilon(s,t)$ evolves according to the following integral dynamics:}

$$e(s,t) = \int_s^t \mathbf{R}(r) dr$$

\textit{proof.} Let $\{x_r\}_{r\in[s,t]}$ follow the marginal flow
\begin{equation}
\frac{d x_r}{dr} = u_r(x_r).
\end{equation}
For the ideal consistency mapping, $f^*(x_t,s,t)=x_s$ for all $t$, hence
\begin{equation}
\frac{d}{dt} f^*(x_t,s,t)
= \partial_t f^* + \nabla_{x_t} f^* \cdot u_t
= 0.
\end{equation}
Taking the total derivative of $e(s,t)$ along the trajectory yields
\begin{equation}
\begin{aligned}
\frac{d}{dt} e(s,t)
&= \frac{d}{dt} f_\theta(x_t,s,t) \\
&= \partial_t f_\theta + \nabla_{x_t} f_\theta \cdot \frac{dx_t}{dt} \\
&= \partial_t f_\theta + \nabla_{x_t} f_\theta \cdot u_t \\
&=: \mathbf{R}(t).
\end{aligned}
\end{equation}
Using the boundary condition $f_\theta(x_s,s,s)=f^*(x_s,s,s)=x_s$, we have $e(s,s)=0$. Integrating from $s$ to $t$ gives
\begin{equation}
e(s,t) - e(s,s) = \int_s^t \mathbf{R}(r)\,dr \implies e(s,t) = \int_s^t \mathbf{R}(r)\,dr. 
\end{equation}

\subsection{Detailed Derivation of the Eq~\eqref{eq:eq7}.}

We aim to show that optimizing Eq.~\eqref{eq:eq7} is equivalent to optimizing Eq.~\eqref{eq:eq6}.
Let $\mathbf{D}(u_t) := \nabla_{x_t} F_\theta(x_t,s,t)\, u_t + \partial_t F_\theta(x_t,s,t)$, and denote $\Delta t = t - s$.

Starting from Eq.~\eqref{eq:eq6}, we replace the marginal velocity $u_t$ in the first term with
the conditional velocity $v_t$, yielding the auxiliary objective
\begin{equation}
\begin{aligned}
\mathcal{L}_v
&= \mathbb{E}_{x_t, v_t} \Big[ \big\| (F_\theta(x_t,s,t) - v_t) + \Delta t\, \mathbf{D}(u_t) \big\|^2 \Big].\\
&= \mathbb{E}_{x_t,v_t}\big[\|F_\theta - v_t\|^2\big]
+ 2 \Delta t\, \mathbb{E}_{x_t,v_t}\big[(F_\theta - v_t)^\top \mathbf{D}(u_t)\big]
+ \Delta t^2\, \mathbb{E}_{x_t}\big[\|\mathbf{D}(u_t)\|^2\big].
\end{aligned}
\end{equation}

Using the law of total expectation and the identity
$u_t = \mathbb{E}[v_t \mid x_t]$, the cross term becomes
\begin{equation}
\begin{aligned}
\mathbb{E}_{x_t,v_t}\big[(F_\theta - v_t)^\top \mathbf{D}(u_t)\big]
&= \mathbb{E}_{x_t} \left[ \mathbb{E}_{v_t|x_t} [ (F_\theta - v_t)^\top \mathbf{D}(u_t) ] \right] \\
&= \mathbb{E}_{x_t} \left[ (F_\theta - \mathbb{E}[v_t | x_t])^\top \mathbf{D}(u_t)\right] \\
&=\mathbb{E}_{x_t}\big[(F_\theta - u_t)^\top \mathbf{D}(u_t)\big].
\end{aligned}
\end{equation}
Moreover, the first term admits the variance decomposition
\begin{equation}
\begin{aligned} \mathbb{E}_{x_t,v_t}[\| F_\theta - v_t \|^2] 
&= \mathbb{E}_{x_t,v_t}[\| F_\theta - (u_t + \delta_t) \|^2] \\ 
&= \mathbb{E}_{x_t,v_t}[\| F_\theta - u_t \|^2] - \mathbb{E}_{x_t,v_t}[2(F_\theta - u_t)^\top \delta_t] + \mathbb{E}_{x_t,v_t}[\| \delta_t \|^2]  \\
&= \mathbb{E}_{x_t,v_t}[\| F_\theta - u_t \|^2] - \mathbb{E}_{x_t}[2(F_\theta - u_t)^\top (u_t-\mathbb{E}[v_t|x_t])] + \mathbb{E}_{x_t,v_t}[\| \delta_t \|^2]  \\
&=\mathbb{E}_{x_t,v_t}\big[\|F_\theta - u_t\|^2\big] + \mathbb{E}_{x_t,v_t}\big[\|u_t - v_t\|^2\big],
\end{aligned}
\end{equation}
where $\mathbb{E}[\|u_t - v_t\|^2] = \mathrm{Tr}(\Sigma_t)$ depends only on the conditional
velocity variance and is independent of the network parameters $\theta$. Therefore, Equation (25) can be reformulated as:
\begin{equation}
\begin{aligned}
\mathcal{L}_v
&= \mathbb{E}_{x_t, v_t}\Big[\big\|(F_\theta(x_t,s,t) - v_t) + \Delta t\, \mathbf{D}(u_t) \big \|^2 \Big].\\
&= \mathbb{E}_{x_t,v_t}\big[\|F_\theta - v_t\|^2\big]
+ 2 \Delta t\, \mathbb{E}_{x_t,v_t}\big[(F_\theta - v_t)^\top \mathbf{D}(u_t)\big]
+ \Delta t^2\, \mathbb{E}_{x_t}\big[\|\mathbf{D}(u_t)\|^2\big].\\
&= \mathbb{E}_{x_t}\big[\|F_\theta - u_t\|^2\big]
+ 2 \Delta t\, \mathbb{E}_{x_t}\big[(F_\theta - u_t)^\top \mathbf{D}(u_t)\big]
+ \Delta t^2\, \mathbb{E}_{x_t}\big[\|\mathbf{D}(u_t)\|^2\big] + \mathrm{Tr}(\Sigma_t).\\
&= \mathbb{E}_{x_t}\Big[\big\|(F_\theta(x_t,s,t) - u_t) + \Delta t\, \mathbf{D}(u_t) \big \|^2 \Big]+C.
\end{aligned}
\end{equation}

Consequently, up to additive constants independent of $\theta$, $\mathcal{L}_v$ is equivalent to the objective in Eq.~\eqref{eq:eq6}.

However, Equation (28) precisely reduces to the Flow Matching objective when $s=t$, which implies that $F_\theta(x_t, t, t)$ can serve as an estimator for the marginal velocity field, i.e., $u_\theta = F_\theta(x_t, t, t)$. By substituting $u_\theta$ into Equation (29), we obtain the following expression:
\begin{equation}
\mathcal{L}_v=\mathbb{E}_{x_t} [ \| (F_\theta(x_t,s,t) - v_t) - (t-s)(\nabla_{x_t} F_\theta \cdot u_\theta + \partial_t F_\theta) \|^2 ] 
\end{equation}

\section{Implementation Details}
\label{details}
We summarize the configurations and hyperparameters of our method in Tab.~\ref{tab:configs}. 
For training, we adopt the standard DiT~\cite{dit} with $2{\times}2$ patches as the model architecture. 
Given that this architecture natively accepts three inputs ($x_t, t, c$), we incorporate two additional two-layer MLPs to embed the time step $s$ and CFG scale $\omega$. 
During the training of FlowConsist-XL/2, we initialize the weights using pre-trained SiT-XL/2 with REPA~\cite{repa} to provide a robust initialization for the marginal velocity. 
For evaluation, we sample 50 images per class to calculate FID.

\begin{table}[ht]
\small
\centering
\caption{\textbf{Configurations for ImageNet $256\times256$.} Left: Model architectural specifications for B/2 and XL/2 variants. Right: Training hyperparameters and loss settings common to both models.}
\label{tab:configs}
\vspace{8pt}

% --- 左侧子表：Model Specs ---
\begin{minipage}[t]{0.48\textwidth}
    \centering
    \renewcommand{\arraystretch}{1.10}
    \begin{tabular}{lcc}
    \toprule
    \textbf{Model} & \textbf{FlowConsit-B/2} & \textbf{FlowConsit-XL/2} \\
    \midrule
    Params (M) & 131 & 676 \\
    FLOPs (G)  & 5.6 & 119.0 \\
    Depth      & 12  & 28 \\
    Hidden dim & 768 & 1152 \\
    Heads      & 12  & 16 \\
    Patch size & $4{\times}4$ & $2{\times}2$ \\
    \midrule
    CFG $\omega$ range & [1, 4] & [1, 4] \\ 
    CFG cond drop & 0.1 & 0.1 \\
    CFG interval  & [0.0, 0.75] & [0.0, 0.75] \\
    \addlinespace[1.4pt] % 微调间距以对齐右表
    \bottomrule
    \end{tabular}
\end{minipage}
\hfill 
% --- 右侧子表：Training Specs ---
\begin{minipage}[t]{0.48\textwidth}
    \centering
    \begin{tabular}{lc}
    \toprule
    \textbf{Training Setup} & \textbf{Value} \\
    \midrule
    Epochs (B/2, XL/2) & 80, 200 \\
    Batch size & 256 \\
    Optimizer & Adam \cite{kingma2014adam} \\
    Learning rate & $1\times10^{-5}$ \\
    Adam $(\beta_1, \beta_2)$ & (0.9, 0.95) \\
    EMA decay & 0.9999 \\
    \midrule
    $(s, t)$ sampler & lognorm($-0.4, 1$) \\
    Ratio of $s=t$ & 60\% \\
    Adaptive weight $p$ & 1.0 \\
    Weight decay & 0.0 \\
    \bottomrule
    \end{tabular}
\end{minipage}
\end{table}

\section{Discussion About Related Work}
\label{discuss}
Recently, a concurrent work, iMF~\cite{geng2025improved}, proposed an implementation scheme similar to our Eq.~\eqref{eq:eq9}. 
iMF reparameterizes the loss of the mean velocity $u_{\text{avg}}$ from MeanFlow into a loss for instantaneous velocity $v$, thereby transforming the training objective into a standard regression form. 
Its training objective can be expressed as $\mathbb{E}_{t,s,x,\epsilon} \|V_\theta - (\epsilon - x)\|$, where $V_\theta = F_\theta(x_t, s, t) + dF_\theta/dt$. 
In the original MeanFlow, the term $dF_\theta/dt = \nabla_{x_t} F_\theta \cdot v_t + \partial_t F_\theta$ introduces a dependency of $V_\theta$ on $v_t = \epsilon - x$, causing $V_\theta(u_\theta, \epsilon - x)$ to violate the standard regression format. 
To rectify this, iMF replaces $v_t$ with the model’s own predicted marginal velocity $u_\theta$.

Despite surface-level implementation similarities, our work and iMF address fundamentally different problems.
iMF reparameterizes the MeanFlow objective into a standard regression form by replacing the conditional velocity with the model’s own predicted marginal velocity, which can be viewed as an effective engineering fix. 
In contrast, our work identifies and formalizes a previously overlooked theoretical issue shared by fast-flow models: \textbf{the use of conditional velocity as a surrogate for marginal velocity introduces systematic trajectory perturbation during training.}

From the perspective of trajectory consistency, we analyze why the practice in Flow Matching (using conditional velocity to approximate marginal velocity) cannot be directly generalized to fast-flow models. 
Although the expectation over all conditional velocities recovers the marginal velocity, treating conditional velocity as an instantaneous velocity estimate causes the optimization target to \textbf{deviate from any single marginal trajectory}. 
As a result, the model is implicitly trained to approximate an average over multiple distinct trajectories, leading to suboptimal convergence and degraded performance in fast-flow settings.

Previous methods, including Consistency Models (CM) and MeanFlow, implicitly rely on this substitution without explicitly distinguishing between marginal and conditional trajectories. 
CM introduced the concept of trajectory consistency but did not formalize the distinction between marginal paths and conditional paths, which was later clarified by Flow Matching.
MeanFlow explicitly defines marginal and conditional velocities, yet assumes that conditional velocity can serve as an equivalent substitute for marginal velocity during training, mirroring the practice in Flow Matching.

Our contribution goes beyond specific implementations.
We provide a \textbf{rigorous theoretical analysis and complete mathematical derivation} showing that conditional-velocity-based training induces a systematic shift in the optimization trajectory of fast-flow models. 
This analysis explains why such models tend to learn an average over multiple trajectories rather than a single consistent marginal trajectory, which is entirely absent from iMF and prior work. 
Therefore, despite superficial similarities in implementation, our work makes a distinct and foundational contribution to the training theory of fast-flow generative models.

\section{More Visual Results}
\label{visual}
\begin{figure}[htbp]
    \centering
    
    \begin{subfigure}[b]{0.8\textwidth}
        \centering
        \includegraphics[width=\textwidth]{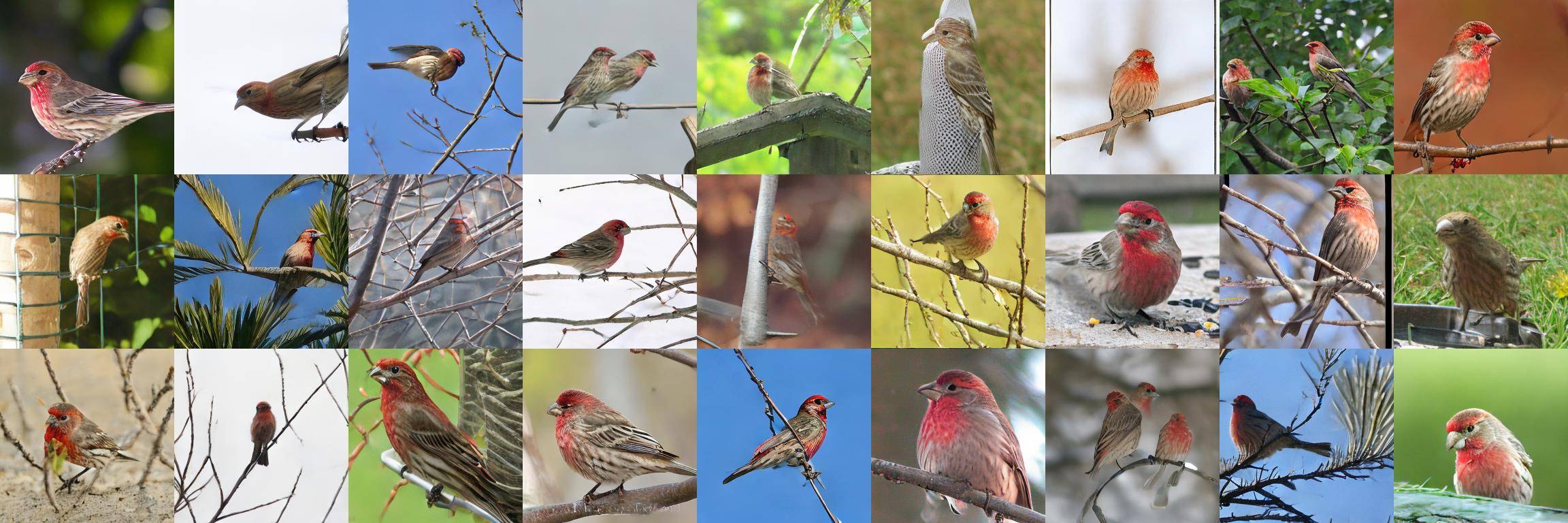}
        \caption*{class 12 (house finch, linnet, Carpodacus mexicanus)}
    \end{subfigure}
    \hfill 
    
    \begin{subfigure}[b]{0.8\textwidth}
        \centering
        \includegraphics[width=\textwidth]{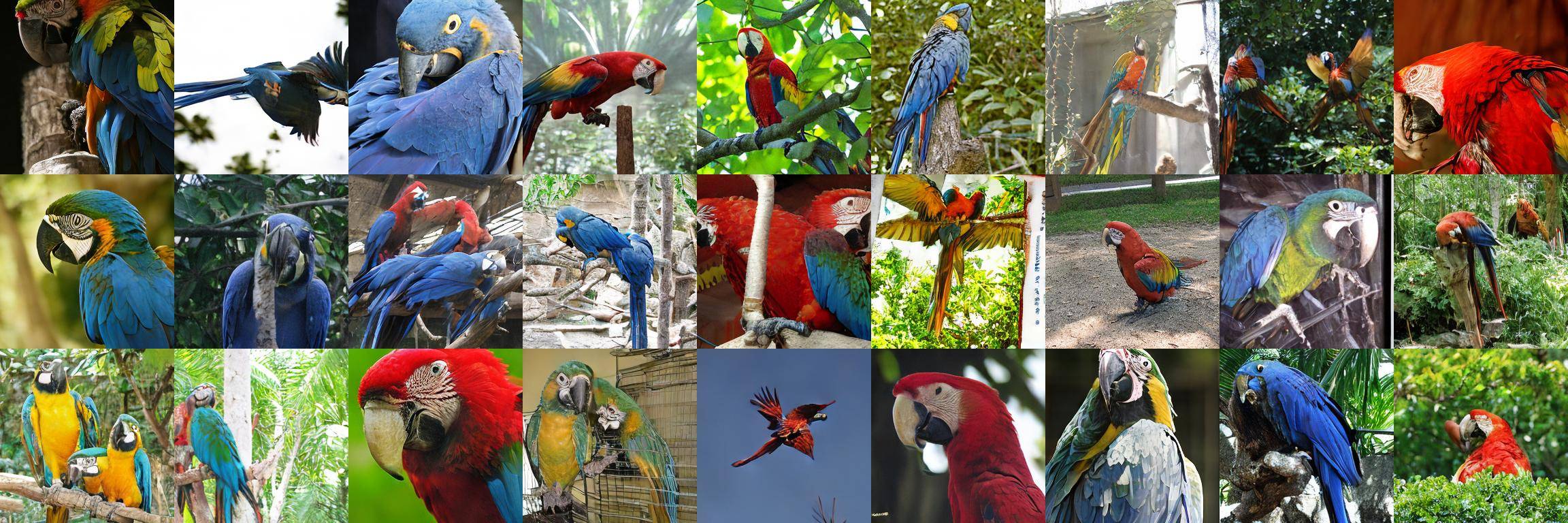}
        \caption*{class 88 (macaw)}
    \end{subfigure}
    \hfill
    
    \begin{subfigure}[b]{0.8\textwidth}
        \centering
        \includegraphics[width=\textwidth]{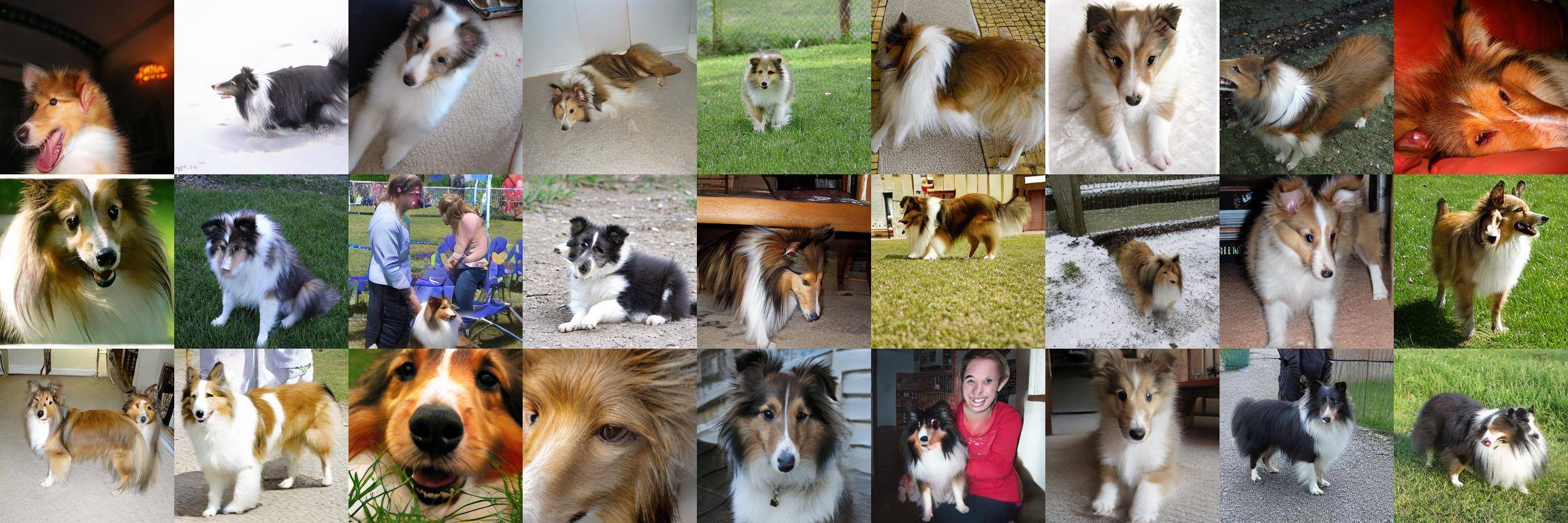}
        \caption*{class 230 (old English sheepdog)}
    \end{subfigure}
    
    \begin{subfigure}[b]{0.8\textwidth}
        \centering
        \includegraphics[width=\textwidth]{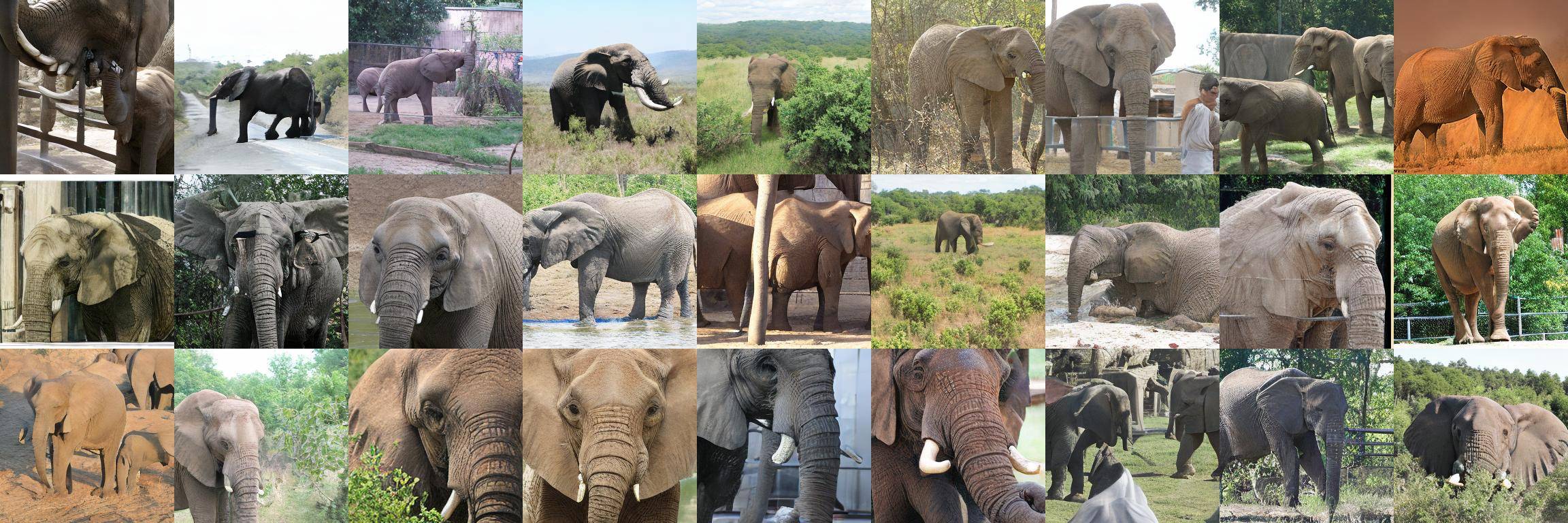}
        \caption*{class 386 (African elephant)}
    \end{subfigure}
    \caption{Uncurated 1-NFE class-conditional generation samples of FlowConsist-XL/2 on ImageNet 256×256}
    \label{fig: visual_1}
\end{figure}

\begin{figure}[htbp]
    \centering
    
    \begin{subfigure}[b]{0.8\textwidth}
        \centering
        \includegraphics[width=\textwidth]{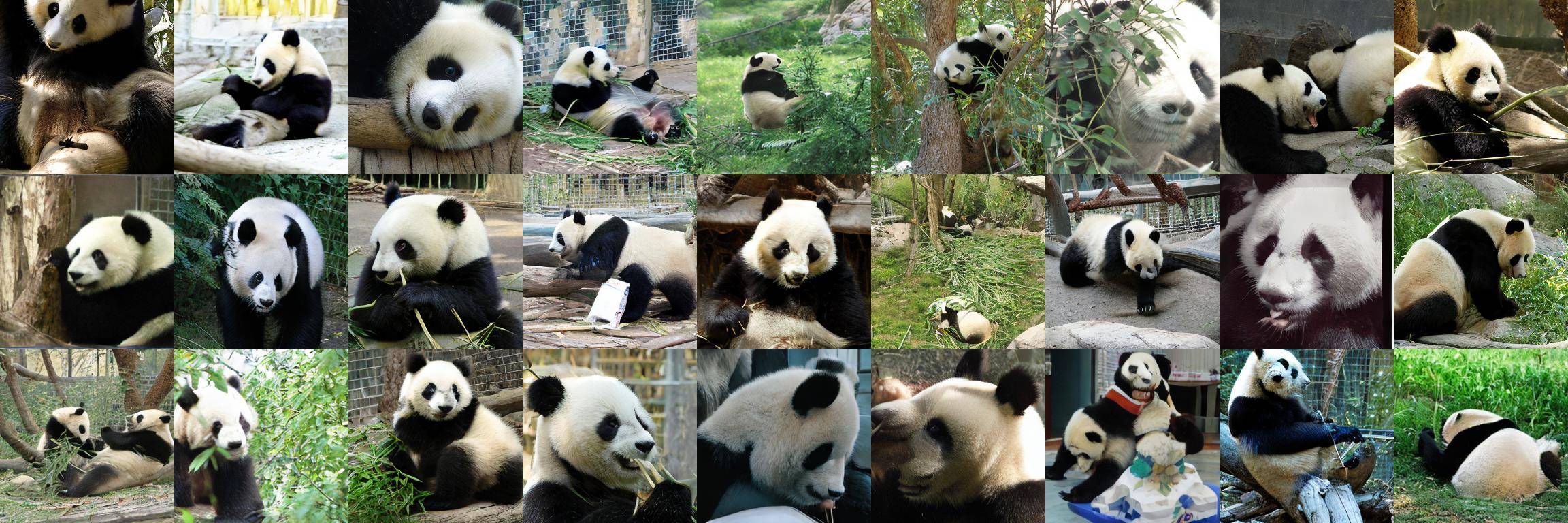}
        \caption*{class 388 (giant panda)}
    \end{subfigure}
    \hfill 
    
    \begin{subfigure}[b]{0.8\textwidth}
        \centering
        \includegraphics[width=\textwidth]{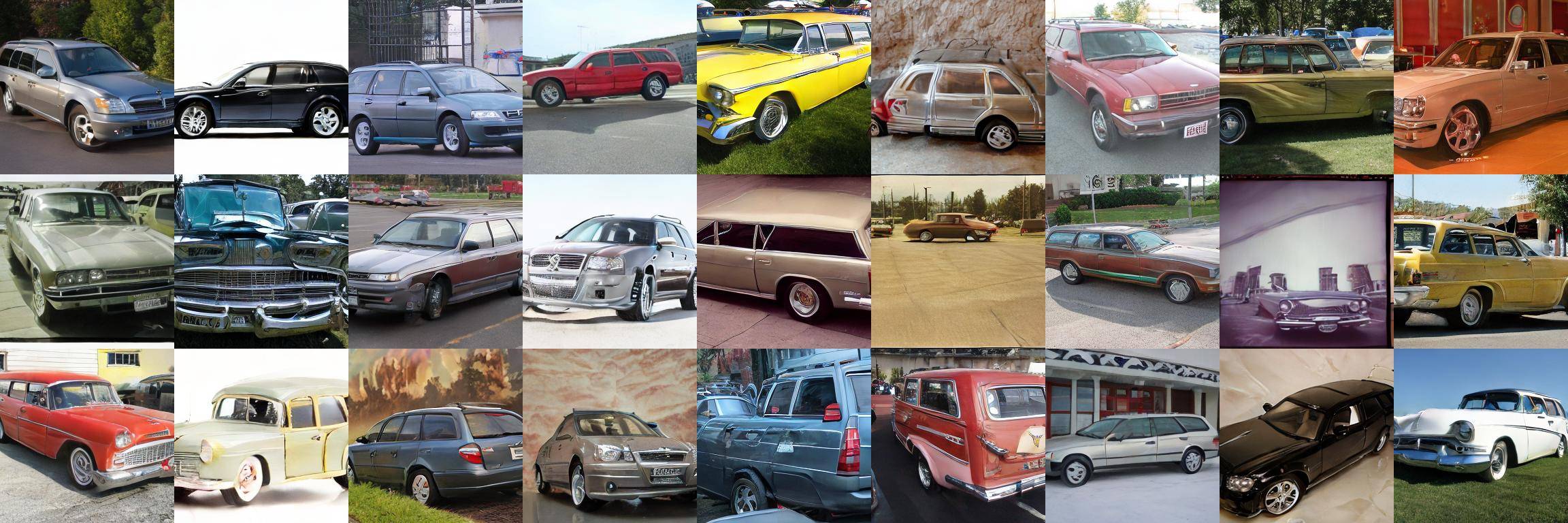}
        \caption*{class 436 (beach wagon, station wagon)}
    \end{subfigure}
    \hfill
    
    \begin{subfigure}[b]{0.8\textwidth}
        \centering
        \includegraphics[width=\textwidth]{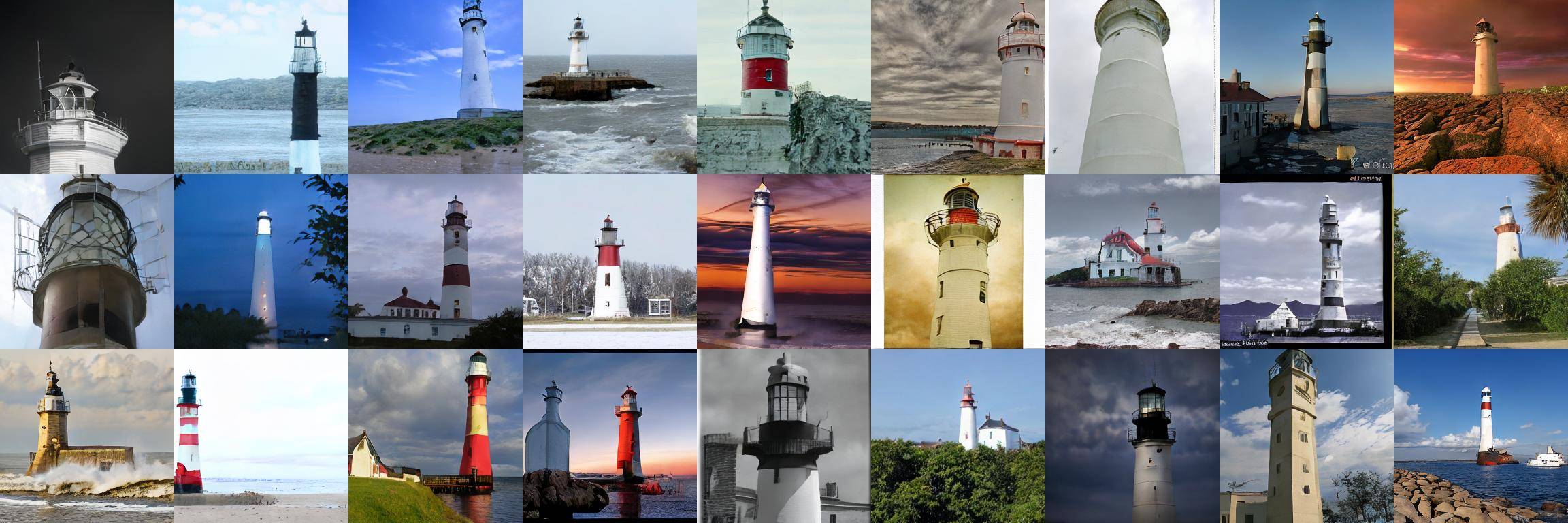}
        \caption*{class 437 (beacon, lighthouse, beacon light, pharos)}
    \end{subfigure}
    
    \begin{subfigure}[b]{0.8\textwidth}
        \centering
        \includegraphics[width=\textwidth]{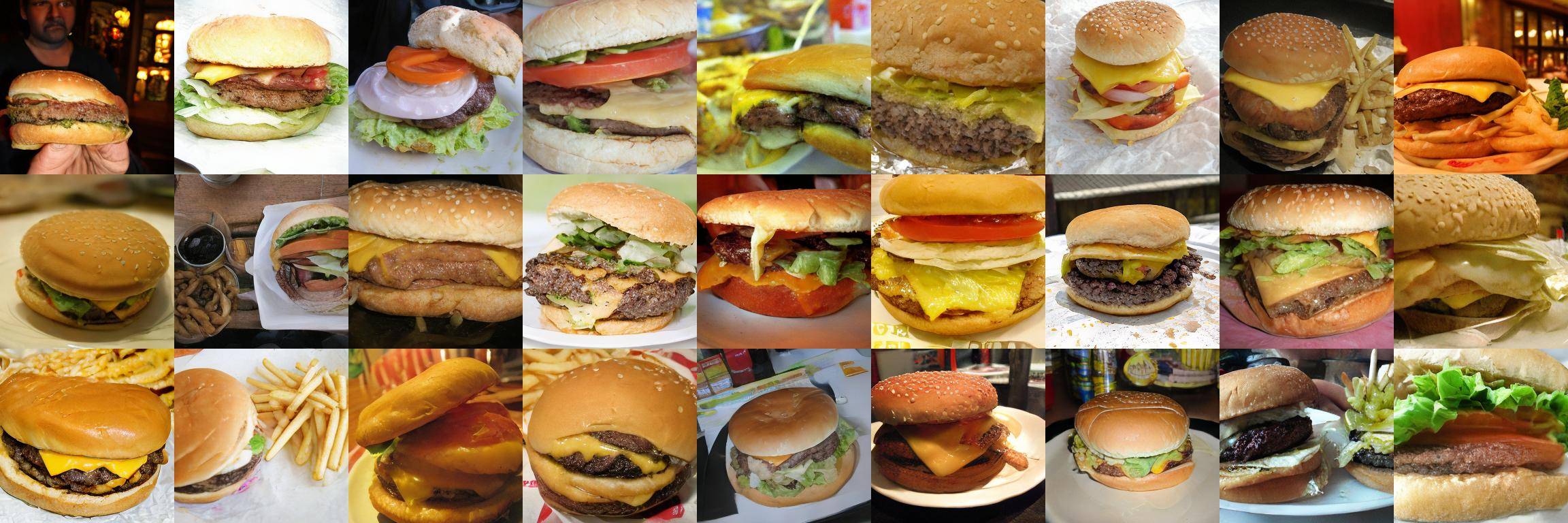}
        \caption*{class 933 (cheeseburger)}
    \end{subfigure}
    \caption{Uncurated 1-NFE class-conditional generation samples of FlowConsist-XL/2 on ImageNet 256×256}
    \label{fig: visual_2}
\end{figure}

We present additional uncurated samples generated by FlowConist-XL/2 at $256\times256$ resolution with 1-NFE in Fig.~\ref{fig: visual_1} and Fig~\ref{fig: visual_2}.

%%%%%%%%%%%%%%%%%%%%%%%%%%%%%%%%%%%%%%%%%%%%%%%%%%%%%%%%%%%%%%%%%%%%%%%%%%%%%%%
%%%%%%%%%%%%%%%%%%%%%%%%%%%%%%%%%%%%%%%%%%%%%%%%%%%%%%%%%%%%%%%%%%%%%%%%%%%%%%%

\end{document}